\documentclass[letterpaper, oneside, 12pt]{article}
\usepackage{amsmath}
\usepackage{graphicx}%
\usepackage{amsfonts}%
\usepackage{amssymb}
\usepackage{overpic}
\usepackage{subfigure}
\usepackage{rotating}
\usepackage{epstopdf}
\usepackage{url}
\usepackage{color}
\usepackage{comment}
\usepackage{enumerate}
\usepackage{mathtools}
\usepackage{multirow}
\usepackage{pgfplots}
\usepackage{pst-solides3d}
\usepackage{algorithm}
\usepackage{algorithmic}
\usepackage{hyperref}

\DeclarePairedDelimiter{\ceil}{\lceil}{\rceil}

\usepackage{setspace}
\usepackage[hmargin=1.0in,vmargin=1.0in]{geometry}

\newcommand{\bA}{ {\boldsymbol A} }

\newcommand{\bD}{ {\boldsymbol D} }

\newcommand{\bH}{ {\boldsymbol H} }

\newcommand{\bI}{ {\boldsymbol I} }

\newcommand{\bk}{ {\boldsymbol k} }
\newcommand{\bK}{ {\boldsymbol K} }

\newcommand{\bL}{ {\boldsymbol L} }
\newcommand{\bm}{ {\boldsymbol m} }

\newcommand{\bs}{ {\boldsymbol s} }
\newcommand{\bS}{ {\boldsymbol S} }

\newcommand{\bx}{ {\boldsymbol x} }
\newcommand{\bX}{ {\boldsymbol X} }
\newcommand{\by}{ {\boldsymbol y} }

\newcommand{\bz}{ {\boldsymbol z} }

\newcommand{\bbeta}{ {\boldsymbol \beta} }

\newcommand{\bgamma}{ {\boldsymbol \gamma} }

\newcommand{\bdelta}{ {\boldsymbol \delta} }
\newcommand{\bDelta}{ {\boldsymbol \Delta} }

\newcommand{\bPhi}{ {\boldsymbol \Phi} }

\newcommand{\bmu}{ {\boldsymbol \mu} }

\newcommand{\bSigma}{ {\boldsymbol \Sigma} }

\newcommand{\bPsi}{ {\boldsymbol \Psi} }

\newcommand{\bzero}{ {\boldsymbol 0} }

\newcommand{\given}{\,|\,}

\newtheorem{theorem}{Theorem}[section]

\newcommand{\qed}{\nobreak \ifvmode \relax \else
      \ifdim\lastskip<1.5em \hskip-\lastskip
      \hskip1.5em plus0em minus0.5em \fi \nobreak
      \vrule height0.75em width0.5em depth0.25em\fi}
\title{Compressed Gaussian Process}
\author{Rajarshi Guhaniyogi and David B. Dunson}
\begin{document}
\maketitle
\begin{abstract}
Nonparametric regression for massive numbers of samples ($n$) and features ($p$) is an increasingly important problem.
In big $n$ settings, a common strategy is to partition the feature space, and then separately apply simple models to each partition set.
We propose an alternative approach, which avoids such partitioning and the associated sensitivity to neighborhood choice and distance
metrics, by using random compression combined with Gaussian process regression.  The proposed approach is particularly motivated by
the setting in which the response is conditionally independent of the features given the projection to a low dimensional
manifold.  Conditionally on the random compression matrix and a smoothness parameter, the posterior distribution for the regression surface and posterior predictive
distributions are available analytically.  Running the analysis in parallel for many random compression matrices and smoothness parameters, model averaging is used to combine the results.  The algorithm can be implemented rapidly even in very big $n$ and $p$ problems, has strong theoretical justification,
and is  found to yield state of the art predictive performance.
\end{abstract}



\section{Introduction}
With recent technological progress, it is now routine in many disciplines to collect data containing massive numbers of features, ranging from thousands to millions or more. To account for complex nonlinear relationships between the features and the response, nonparametric regression models are employed. For example,
\begin{align*}
y =\mu_0(\bx)+\epsilon,\:\:\epsilon\sim N(0,\sigma^2),
\end{align*}
where $\bx\in \mathcal{R}^p$, $\mu_0(\cdot)$ is the unknown regression function and $\epsilon$ is a residual. When $p$ is massive, estimating $\mu_0$ can lead to a statistical and computational curse of dimensionality.  One strategy for combatting this curse is dimensionality reduction via variable selection or (more broadly) subspace learning, with the high-dimensional features replaced with their projection to a $d$-dimensional subspace or manifold with $d \ll p$.  In many applications, the relevant information about the high-dimensional features can be encoded in such low dimensional coordinates.

There is a vast frequentist literature on subspace learning for regression, typically employing a two stage approach. In the first stage, a dimensionality reduction technique is used to obtain lower dimensional features that can ``faithfully'' represent the higher dimensional features. Examples include principal components analysis and more elaborate methods that accommodate non-linear subspaces, such as  isomap (\cite{tenenbaum2000global}) and Laplacian eigenmaps (\cite{belkin2003laplacian,guerrero2011laplacian}). These techniques rely on eigen-decomposition of an $n\times n$ matrix, making efficient computation in large $n$ challenging.  There is a rich literature focused on freeing this bottleneck. For example,
  \cite{talwalkar2008large} employs a column sampling algorithm that only requires eigen-decomposition of an $m\times m$ matrix without losing much accuracy, even with $m\ll n$. Once lower dimensional features are obtained, the second stage uses these features in standard regression and classification procedures as if they were observed initially. Such two stage approaches rely on learning the manifold structure embedded in the high dimensional features, which adds unnecessary computational burden when inferential interest lies mainly in prediction.

Another thread of research focuses on prediction using divide-and-conquer techniques. As the number of features increases, the problem of finding the best splitting attribute becomes intractable, so that CART (\cite{breiman1984classification}), MARS and multiple tree models, such as Random Forest (\cite{breiman2001random}) cannot be efficiently applied. A much simpler approach is to apply high dimensional clustering techniques, such as metis, cover trees and spectral clustering. Once the observations are clustered into a few groups, simple models (glm, Lasso etc) are fitted in each cluster.  Such methods are sensitive to clustering, do not characterize predictive uncertainty, and may lack efficiency, an important consideration outside the $n \gg p$ setting.  There is also a recent literature on scaling up sparse optimization methods, such as Lasso, to huge $n$ and $p$ settings relying on algorithms that can exploit multiple processors in a distributed manner (\cite{boyd2011distributed}).  However, such methods are yet to be developed for non-linear manifold regression, which is the central focus of this article.


 This naturally motivates Bayesian models that simultaneously learn the mapping to the lower-dimensional subspace along with the regression function in the coordinates on this subspace, providing a characterization of predictive uncertainties. \cite{tokdar2010bayesian} proposes a logistic Gaussian process approach, while \cite{reich2011sufficient} use finite mixture models for
 sufficient dimension reduction. \cite{page2013classification} propose a Bayesian nonparametric model for learning of an affine subspace in classification problems.  These approaches have the disadvantages of being limited to linear subspaces, lacking scalability beyond a few dozen features and having potential sensitivity to features corrupted with noise.  There is also a literature on Bayesian methods that accommodate non-linear subspaces, ranging from Gaussian process latent variable models (GP-LVMs) (\cite{lawrence2005probabilistic}) for probabilistic nonlinear PCA to mixture factor models (\cite{chen2010compressive}).  However, such methods similarly face barriers in scaling up to large $n$ and/or $p$.  There is a heavy computational price for learning the number of latent variables, the distribution of the latent variables, and the mapping functions while maintaining identifiability restrictions.

Recently, \cite{yang2013bayesian} show that this computational burden can be largely bypassed by using usual Gaussian process (GP) regression without attempting to learn the mapping to the lower-dimensional subspace.  They showed that when the features lie on a $d$-dimensional manifold embedded in the $p$-dimensional feature space with $d \ll p$ and the regression function is not highly smooth, the optimal rate can be obtained using GP regression with a squared exponential covariance in the original high-dimensional feature space.  This is an exciting theoretical result, which provides motivation for the approach in this article, which is focused on scalable Bayesian nonparametric regression in large $p$ and $n$ settings. For broader applicability than (\cite{yang2013bayesian}), we accommodate features that are contaminated by noise and hence do not lie exactly on a low-dimensional manifold.  In addition, we facilitate massive scaling in both $p$ and $n$ by bypassing MCMC and reducing matrix inversion bottlenecks via random projections. Sensitivity to the random projection and to tuning parameters is eliminated through the use of Bayesian model averaging. To our knowledge, no Bayesian manifold regression technique has yet been developed that can scale up for large sample size and massive number of features yielding accurate predictive inference rapidly.

Section 2 proposes the model and computational approach in large $p$ settings.  Section 3 describes extensions to large $n$, and section 4 develops theoretical justification.  Section 5 contains simulation examples relative to state-of-the-art competitors.  Section 6 presents an image data application, Section 7 concludes the paper with a discussion.

\section{Compressed Gaussian process regression}

\subsection{Model}
For subjects $i=1,\ldots,n$, let $y_i \in \mathcal{Y}$ denote a response with associated  features $\bx_i = (x_{i1},\ldots,x_{ip})'=(z_{i1},\ldots,z_{ip})'+
(\delta_{i1},\ldots,\delta_{ip})'=\bz_i+\bdelta_i$, $\bz_i\in\mathcal{M}$, $\bdelta_i\in\mathcal{R}^p$, where $\mathcal{M}$ is a $d$-dimensional manifold embedded in the ambient space $\mathcal{R}^p$.  We assume that the response $y\in \mathcal{Y}$ is continuous. The measured features do not fall exactly on the manifold $\mathcal{M}$ but are corrupted by noise.  We assume a compressed nonparametric regression model
\begin{equation}\label{eq:Bayessquash}
y_i =\mu\big(\bPsi\bx_i\big)+\epsilon_i,\quad \epsilon_i \sim N(0, \sigma^2),
\end{equation}
with the residuals modeled as Gaussian with variance $\sigma^2$, though other distributions including heavy-tailed ones can be accommodated.  $\bPsi$ is an $m\times p$ matrix that compresses $p$-dimensional features to dimension $m$.  Following a Bayesian approach, we choose a prior distribution for the regression function $\mu$ and residual variance $\sigma^2$, while randomly generating $\bPsi$ following precedence in the literature on feature compression (\cite{maillard2009compressed,fard2012compressed,guhaniyogi2013bayesian}).  These earlier approaches differ from ours in focusing on parametric regression.  Specifically, we independently draw elements $\{ \Psi_{ij} \}$ of $\bPsi$ from $N(0,1)$, and then normalize the rows using Gram-Schmidt orthogonalization.

We assume that $\mu \in \mathcal{H}_s$ is a continuous function belonging to $\mathcal{H}_s$, a Holder class with smoothness $s$.  To allow $\mu$ to be unknown, we use a Gaussian process (GP) prior, $\mu \sim \mbox{GP}( 0, \sigma^2 \kappa)$ with the covariance function chosen to be squared exponential
\begin{equation}\label{eq:sqexp}
\kappa(\bx_i,\bx_j;\lambda)=\exp\left(-\lambda||\bx_i-\bx_j||^2\right),
\end{equation}
with $\lambda$ a smoothness parameter and $||\cdot||^2$ the Euclidean norm.  To additionally allow the residual variance $\sigma^2$ and smoothness $\lambda$ to be unknown, we let
\begin{align*}
\sigma^2\sim IG(a,b),\quad \lambda^d\sim Ga(a_0,b_0),
\end{align*}
with $IG()$ and $Ga()$ denoting the inverse-gamma and gamma densities, respectively. 
The powered gamma prior for $\lambda$ is motivated by the result of (\cite{van2009adaptive}) showing minimax adaptive rates of $n^{-s/(2s+p)}$
for a GP prior with squared exponential covariance and powered gamma prior.  This is the optimal rate for nonparametric regression in the original $p$-dimensional ambient space.  The rate can be improved to $n^{-s/(2s+d)}$ when $\bx_i \in \mathcal{M}$, with $\mathcal{M}$ a $d$-dimensional manifold. \cite{yang2013bayesian} shows that a GP prior with powered gamma prior on the smoothness can achieve this rate.

In many applications, features may not lie exactly on $\mathcal{M}$ due to noise and corruption in the data.  We apply random compression in (\ref{eq:Bayessquash}) to de-noise the features, obtaining $\bPsi\bx_i$ much more concentrated near a lower-dimensional subspace than the original $\bx_i$.  With this enhanced concentration, the theory in \cite{yang2013bayesian} suggests excellent performance for an appropriate GP prior.  In addition to de-noising, this approach has the major advantage of bypassing estimation of a geodesic distance along the unknown manifold $\mathcal{M}$ between any two data points $\bx_i$ and $\bx_{i'}$.

\subsection{Posterior form}
Let $\bmu=(\mu(\bPsi\bx_1),...,\mu(\bPsi\bx_n))'$ and $\bK_1=(\kappa(\bPsi\bx_i,\bPsi\bx_j;\lambda))_{i,j=1}^{n}$.
The  prior distribution on $\mu, \sigma^2$ induces a normal-inverse gamma (NIG) prior  on $(\bmu,\sigma^2)$,
\begin{equation*}
(\bmu\given\sigma^2) \sim N(\bzero,\sigma^2\bK_1),\:\sigma^2\sim IG(a,b),
\end{equation*}
leading to a NIG posterior distribution for $(\bmu,\sigma^2)$ given $y, \bPsi\bx,\lambda$.  In the special case in which $a,b \to 0$, we obtain Jeffrey's prior and the posterior distribution is
\begin{align}\label{eq:postdist}
\bmu\given\by &\sim t_{n}(\bm,\bSigma)\\
\sigma^2\given\by &\sim IG(a_1,b_1),
\end{align}
where $a_1=n/2$, $b_1=\by'\left(\bK_1+\bI\right)^{-1}\by/2$,
$\bm=\left[\bI+\bK_1^{-1}\right]^{-1}\by$, $\bSigma=(2b_1/n)\left[\bI+\bK_1^{-1}\right]^{-1}$,
and $t_{\nu}(\bm,\bSigma)$ denotes a multivariate-$t$ distribution with $\nu$ degrees of freedom, mean $\bm$ and covariance $\bSigma$.

Hence, the exact posterior distribution of $(\bmu,\sigma^2)$ conditionally on $(\bPsi,\lambda)$ is available analytically. The predictive of $\by^*=(y_1^*,...,y_{n_{pred}}^*)'$ given $\bX^*=\left(\bx_{1}^{*'},...,\bx_{n_{pred}}^{*'}\right)'$ and $\bPsi,\lambda$ for new $n_{pred}$ subjects marginalizing out $(\bmu,\sigma^2)$ over their posterior distribution is available analytically as
\begin{equation}\label{eq:fyxPhi}
\by^* | \bx_{1}^*,...,\bx_{n_{pred}}^*, \by\sim t_{n}\left(\mu_{pred},\sigma_{pred}^2\right),
\end{equation}
where $\bK_{pred}=\{ \kappa(\bx_{i}^*,\bx_{j}^*;\lambda)\}_{i,j=1}^{n_{pred}}$, $\bK_{pred,1}=\{\kappa(\bx_{i}^*,\bx_{j};\lambda)\}_{i=1,j=1}^{i=n_{pred},j=n}$,
$\bK_{1,pred}=\bK_{pred,1}'$, $\mu_{pred}=\bK_{pred,1}\left(\bI+\bK_1\right)^{-1}\by$, $\sigma_{pred}^2=(2b_1/n)\left[\bI+\bK_{pred}-\bK_{pred,1}\left\{\bI+\bK_1\right\}^{-1}\bK_{1,pred}\right]$.

\subsection{Model averaging}\label{sec:modavg}

The approach described in the previous section can be used to obtain a posterior distribution for $\bmu$ and a predictive distribution for $\by^*=(y_1^*,...,y_{n_{pred}}^*)$ given $\bX^*$ for a new set of $n_{pred}$ subjects {\em conditionally} on the $m \times p$ random projection matrix $\bPsi$ and the scaling parameter $\lambda$. To accomplish robustness with respect to the choice of $(\bPsi,\lambda)$ and the subspace dimension $m$, following \cite{guhaniyogi2013bayesian}, we propose to generate $s$ random matrices having different $m$ and $s$ different $\lambda$ from $Unif(3/d_{max},3/d_{min})$, 
$(\Psi^{(l)},\lambda^{(l)})$ , and then use model averaging to combine the results. We choose $d_{max}=\max_{i,j=1,..,n}||\bx_i-\bx_j||^2$, $d_{min}=\min_{i,j=1,..,n}||\bx_i-\bx_j||^2$. To make matters more clear, let $\mathcal{M}_l$, $l=1,\ldots,s$, represent (\ref{eq:Bayessquash}) with $m_l$ number of rows. Corresponding to the model $\mathcal{M}_l$, we denote $\bPsi$, $\lambda$, $\bmu$ and $\sigma^2$  by $\bPsi^{(l)}$, $\lambda^{(l)}$, $\bmu^{(l)}$ and $\sigma^{2(l)}$ respectively. Let $\mathcal{M} = \{ \mathcal{M}_1,\ldots, \mathcal{M}_s \}$ denote the set of models corresponding to different random projections, $\mathcal{D} = \{ (y_i, \bx_i), i=1,\ldots,n \}$ denote the observed data, and $\by^*$ denote the data for future subjects with features$\bX^*$.  Then, the predictive density of $\by^*$ given $\bX^*$ is
\begin{equation}\label{eq:modelaver}
f(\by^* | \bX^*, \mathcal{D}) = \sum_{l=1}^s f(\by^* | \bX^*, \mathcal{M}_l, \mathcal{D})P(\mathcal{M}_l\given \mathcal{D}),
\end{equation}
where the predictive density of $\by^*$ given $\bX^*$ under projection $\mathcal{M}_l$ is given in (\ref{eq:fyxPhi})
and the posterior probability weight on projection $\mathcal{M}_l$ is
\begin{equation*}
P(\mathcal{M}_l\given \mathcal{D})=\frac{P(\mathcal{D}\given \mathcal{M}_l)P(\mathcal{M}_l)}{\sum_{h=1}^s P(\mathcal{D}\given \mathcal{M}_h)P(\mathcal{M}_h)}.
\end{equation*}
Assuming equal prior weights for each random projection, $P(\mathcal{M}_l)=1/s$.  In addition, the marginal likelihood under $\mathcal{M}_l$ is
\begin{equation}\label{eq:marg_model}
P(\mathcal{D}\given\mathcal{M}_l)=\int P(\mathcal{D}\given\mathcal{M}_l,\bmu^{(l)},\sigma^{2(l)})\pi(\bmu^{(l)},\sigma^{2(l)}).
\end{equation}
 After a little algebra, one observes that for
(\ref{eq:Bayessquash}) with $(\bmu\given\sigma^2)\sim N(\bzero,\sigma^2\bK_1)$, $\pi(\sigma^2)\propto \frac{1}{\sigma^2}$, $P(\mathcal{D}\given\mathcal{M}_l)$ is
\begin{equation*}
P(\mathcal{D}\given\mathcal{M}_l)=\frac{1}{\left|\bK_1+\bI\right|^{\frac{1}{2}}}
\frac{2^{\frac{n}{2}}\Gamma(\frac{n}{2})}{\left[\by'\left(\bK_1+\bI\right)^{-1}\by\right]^{\frac{n}{2}}(\sqrt{2\pi})^{n}}.
\end{equation*}
Plugging in the above expressions in (\ref{eq:modelaver}), one obtains the posterior predictive distribution as a weighted average of $t$ densities. Given that
the computation over different sets of $\bPsi,\lambda$ are not dependent on each other, the calculations are embarrassingly parallel with a trivial expense for combining.
The main computational expense comes from the inversion of an $n\times n$ matrix under the $l$th random projection. In the next section, we develop approaches for accelerating this inversion for large $n$.

\section{Scaling to large $n$}\label{sec:largen}

Fitting (\ref{eq:Bayessquash}) using model averaging requires computing inverses and determinants of covariance matrices of the order $n\times n$. In problems with large $n$, this adds a heavy computational burden of the order of $O(n^3)$. Additionally, as dimension increases, matrix inversion becomes more unstable with the propagation of errors due to finite machine precision. This problem is further exacerbated if the covariance matrix is nearly rank deficient.

To address such issues, existing solutions rely on approximating $\mu(\cdot)$ by another process $\tilde{\mu}(\cdot)$, which is more tractable computationally.  One popular approach constructs $\tilde{\mu}(\cdot)$ as a finite basis approximation via kernel convolution (\cite{higdon2002space}) or kalman filtering (\cite{wikle1999dimension}). Alternatively, one can let $\tilde{\mu}(\cdot)=\mu(\cdot)\eta(\cdot)$, where $\eta(\cdot)$ is a Gaussian process having compactly supported correlation function that essentially makes the covariance matrix of $\left(\tilde{\mu}(\bx_1),....,\tilde{\mu}(\bx_n)\right)$ sparse (\cite{kaufman2008covariance}), facilitating inversion through efficient sparse solvers.

 \cite{banerjee2008gaussian} proposes a low rank approach that imputes $\mu(\cdot)$ conditionally on a few knot-points, closely related to subset of regressor methods in machine learning (\cite{smola2000sparse}).  Subsequently, \cite{finley2009hierarchical} in statistics and \cite{snelson2006sparse} in machine learning report bias in
parameter estimation for the proposed approaches (i.e. \cite{banerjee2008gaussian,smola2000sparse}) and suggest possible remedies for bias adjustments.  To avoid sensitivity to knot selection in these approaches,
 \cite{banerjee2013efficient} approximates  $\mu(\cdot)$ using $\tilde{\mu}(\cdot)=E[\mu(\cdot)\given\bPhi\mu(\bX)]+\epsilon_{\bPhi}(\cdot)$, with $\bPhi$ an $m \times n$, $m\ll n$ random matrix with $\Phi_{ij}\sim N(0,1)$. $\epsilon_{\bPhi}(\bx)$ are independent feature specific noises with $\epsilon_{\bPhi}(\bx)\sim N(\bzero,\mbox{var}(\mu(\bx))-\mbox{var}(\tilde{\mu}(\bx)))$,
which are introduced for bias correction similar to \cite{finley2009hierarchical}.

We adapt \cite{banerjee2013efficient} from usual GP regression to our compressed manifold regression setting. In particular, let
\begin{align}\label{eq:largen}
y=\tilde{\mu}_{\bPhi}\left(\bPsi\bx\right)+\epsilon_{\bPhi}(\bPsi\bx)+\epsilon,\:\epsilon\sim N(0,\sigma^2),
\end{align}
where $\tilde{\mu}_{\bPhi}(\bPsi\bx)=E[\mu(\bPsi\bx)\given\bPhi\mu(\bX\bPsi')]$, $\epsilon_{\bPhi}(\bPsi\bx)\given\sigma^2\sim N(0,\sigma_{\epsilon}^2(\bx))$,\\
$\sigma_{\epsilon}^2(\bx)=\sigma^2\left[\kappa(\bPsi\bx,\bPsi\bx;\lambda)-(\bPhi\bk_{\bx})'\left\{\bPhi\bK_1\bPhi'\right\}^{-1}(\bPhi\bk_{\bx})\right]$ and\\
$\bk_{\bx}=(\kappa(\bPsi\bx,\bPsi\bx_1;\lambda),....,\kappa(\bPsi\bx,\bPsi\bx_n;\lambda))'$.
Denoting $\bH_1=diag(\bK_1-\bK_1\bPhi'(\bPhi\bK_1\bPhi')^{-1}\bPhi\bK_1)+\bI$ and $\bH_2=\bK_1\bPhi'(\bPhi\bK_1\bPhi')^{-1}\bPhi$, marginal posterior distributions of
$\mu$ and $\sigma^2$ are available in analytical forms
\begin{align*}
\bmu\given\by\sim t_{n}(\bm_{RGP},\bSigma_{RGP}),\:\:\:\sigma^2\given\by\sim IG(a_2,b_2),
\end{align*}
where $a_2=n/2$, $b_2=\by'\left(\bH_1+\bH_2\bK_1\right)^{-1}\by/2$,
$\bm_{RGP}=\left[\bH_2'\bH_1^{-1}\bH_2+\bK_1^{-1}\right]^{-1}\bH_2'\bH_1^{-1}\by$, $\bSigma_{RGP}=(2b_2/n)\left[\bH_2'\bH_1^{-1}\bH_2+\bK_1^{-1}\right]^{-1}$.
Owing to the special structure of $\bSigma_{RGP}$ and $\bm_{RGP}$, $n\times n$ matrix inversion can be efficiently achieved by Sherman-Woodbury-Morrison matrix inversion technique.

Attention now turns to prediction from (\ref{eq:largen}). The predictive of $\by^*=(y_1^*,...,y_{n_{pred}}^*)'$ given $\bX^*=\left(\bx_{1}^{*'},...,\bx_{n_{pred}}^{*'}\right)'$ and $\bPsi,\lambda$ for new $n_{pred}$ subjects marginalizing out $(\bmu,\sigma^2)$ over their posterior distribution is available analytically as
\begin{equation}\label{eq:fyxPhi}
\by^* | \bx_{1}^*,...,\bx_{n_{pred}}^*, \by\sim t_{n}\left(\mu_{pred},\sigma_{pred}^2\right),
\end{equation}
where $\bK_{pred}=\{ \kappa(\bx_{i}^*,\bx_{j}^*;\lambda)\}_{i,j=1}^{n_{pred}}$, $\bK_{pred,1}=\{\kappa(\bx_{i}^*,\bx_{j};\lambda)\}_{i=1,j=1}^{i=n_{pred},j=n}$,
$\bK_{1,pred}=\bK_{pred,1}'$, $\mu_{pred}=\bK_{pred,1}\left(\bI+\bK_1\right)^{-1}\by$, $\sigma_{pred}^2=(2b_1/n)\left[\bI+\bK_{pred}-\bK_{pred,1}\left\{\bI+\bK_1\right\}^{-1}\bK_{1,pred}\right]$.
 Evaluating the above expression
requires inverting matrices of order $m_{\bPhi}\times m_{\bPhi}$. Model averaging is again employed to limit sensitivity over the choices of $\bPsi,\lambda$. Following similar calculations as in section~\ref{sec:modavg}, model averaging weights are found to be
\begin{equation*}
P(\mathcal{D}\given\mathcal{M}_l)=\frac{1}{\left|\bH_2\bK_1+\bH_1\right|^{\frac{1}{2}}}
\frac{2^{\frac{n}{2}}\Gamma(\frac{n}{2})}{\left[\by'\left(\bH_2\bK_1+\bH_1\right)^{-1}\by\right]^{\frac{n}{2}}(\sqrt{2\pi})^{n}}.
\end{equation*}
Model averaging is performed on a wide interval of possible $m$ values determined by the ``compressed sample size" $m_{\bPhi}$ and $p$, analogous to section~\ref{sec:modavg}.

An important question that remains is how much information is lost in compressing the high-dimensional feature vector to a much lower dimension? In particular, one would expect to pay a price for the huge computational gains in terms of predictive performance or other metrics. We address this question in two ways.  First we argue satisfactory theoretical performance in prediction in a large $p$, large $n$ asymptotic paradigm in Section 3. Then, we will consider practical performance in finite samples using simulated and real data sets.

\section{Convergence analysis}\label{sec:convergencerate}

 This section provides theory supporting the excellent practical performance of the proposed method. In our context the feature vector $\bx$ is assumed to be $\bx=\bz+\bdelta$, $\bz\in\mathcal{M}$, $\bdelta\in\mathcal{R}^p$. Compressing the feature vector results in compressing $\bz$ and the noise followed by their addition, $\bPsi\bx=\bPsi\bz+\bPsi\bdelta$. The following two directions are used to argue that compression results in near optimal inference.
 \begin{enumerate}[(A)]
\item When features lie on a manifold a two stage estimation procedure (compression followed by a Gaussian process regression) leads to optimal convergence properties. This is used to show that using $\{\bPsi\bz_i\}_{i=1}^{n}$ as features in the Gaussian process regression yields the optimal rate of convergence.
\item Noise compression through $\bPsi$ mitigates the deleterious effect of noise in $\bx$ on the resulting performance.
 \end{enumerate}

Let $\mu_0(\cdot)$ and $\mu(\cdot)$ be the true and the fitted regression functions respectively.
Define $\rho(\mu,\mu_0)^2=\frac{1}{n}\sum_{i=1}^{n}(\mu(\bx_i)-\mu_0(\bx_i))^2$ as the distance between $\mu$, $\mu_0$ under a fixed design. When the design is random, let $\rho(\mu,\mu_0)^2=\int_{\mathcal{M}}(\mu(\bx)-\mu_0(\bx))^2 F(d\bx)$, where $F$ is the marginal distribution of the features. Denote $\Pi(\cdot|y_1,...,y_n)$ to be the posterior distribution given $y_1,...,y_n$. Then the interest lies in the rate at which the posterior contracts around $\mu_0$ under the metric $\rho(\cdot,\cdot)$. This calls for finding a sequence $\{\zeta_n\}_{n\geq 1}$ of lower bounds such that
\begin{align}\label{eq:cont_rate}
\Pi(\rho(\mu,\mu_0)>\zeta_n\given y_1,...,y_n)\rightarrow 0,\:\mbox{as}\:n\rightarrow\infty.
\end{align}

\textbf{Definition:} Given two manifolds $\mathcal{M}$ and $\mathcal{N}$, a differentiable map $f:\mathcal{M}\rightarrow\mathcal{N}$ is called a diffeomorphism if it is a bijection and its inverse $f^{-1}:\mathcal{N}\rightarrow\mathcal{M}$ is differentiable. If these functions are $r$ times continuously differentiable, $f$ is called a $C^{r}$-diffeomorphism.

Our analysis builds on the following result (Theorem 2.3 in \cite{yang2013bayesian}).
\begin{theorem}\label{th:diffeomorph}
Assume $\mathcal{M}$ is a $d$ dimensional $\mathcal{C}^{r_1}$compact sub-manifold of $\mathcal{R}^p$. Let $G:\mathcal{M}\rightarrow\mathcal{R}^p$ be the embedding map so that $G(\mathcal{M})\simeq\mathcal{M}$. Further assume $T:\mathcal{R}^{p}\rightarrow\mathcal{R}^m$ is a dimensionality reducing map s.t. the restriction $T_{\mathcal{M}}$ of $T$ on $G(\mathcal{M})$ is a $\mathcal{C}^{r_2}$-diffeomorphism onto its image. Then for any $\mu_0\in\mathcal{C}^{s}$ with
$s\leq\min\{2,r_1-1,r_2-1\}$, a Gaussian process prior on $\mu$ with features $\{T(\bz_i)\}_{i=1}^n$, $\bz_i\in\mathcal{M}$, leads to a posterior contraction rate at least
$\zeta_n=n^{-s/(2s+d)}\log(n)^{d+1}$. This is a huge improvement upon the minimax optimal adaptive rate of $n^{-s/(2s+p)}$ without the manifold structure in the features.
\end{theorem}
We use the above result in our context. Define the linear transformation $T(\bz)=\bPsi\bz$.
Using the property of random projection matrix, we have that, given $\kappa\in (0,1)$, if the projected dimension
$m>O(\frac{m}{\kappa^2}\log(Cp\kappa^{-1})\log(\phi_n^{-1}))$ then with probability greater than $1-\phi_n$, the following relationship holds
for every point $\bz_i,\bz_j\in\mathcal{M}$,
\begin{align}\label{eq:JL}
(1-\kappa)\sqrt{\frac{m}{p}}||\bz_i-\bz_j||<||T(\bz_i)-T(\bz_j)||<(1+\kappa)\sqrt{\frac{m}{p}}||\bz_i-\bz_j||,
\end{align}
implying that $T$ is a diffeomorphism onto its image with probability greater than $(1-\phi_n)$. Define $\mathcal{A}_n=\{\mbox{Equation}\:\: \ref{eq:JL}\:\mbox{holds}\}$ so that
$P(\mathcal{A}_n)>1-\phi_n$.
\begin{align*}
\Pi(d(\mu,\mu_0)>\zeta_n|y_1,...,y_n) &=\Pi(d(\mu,\mu_0)>\zeta_n|y_1,...,y_n,\mathcal{A}_n)P(\mathcal{A}_n)\\
&+\Pi(d(\mu,\mu_0)>\zeta_n|y_1,...,y_n,\mathcal{A}_n')P(\mathcal{A}_n')\\
&<\Pi(d(\mu,\mu_0)>\zeta_n|y_1,...,y_n,\mathcal{A}_n)+P(\mathcal{A}_n')\\
&<\Pi(d(\mu,\mu_0)>\zeta_n|y_1,...,y_n,\mathcal{A}_n)+\phi_n.
\end{align*}
On $\mathcal{A}_n$, $T$ is a diffeomorphism. Therefore, Theorem~\ref{th:diffeomorph} implies that with features $\{T(\bz_i)\}_{i=1}^{n}$
$\Pi(d(\mu,\mu_0)>\zeta_n|y_1,...,y_n,\mathcal{A}_n)\rightarrow 0$. Finally, assuming $\phi_n\rightarrow 0$ yields $\Pi(d(\mu,\mu_0)>\zeta_n|y_1,...,y_n)
\rightarrow 0$ with features $\{T(\bz_i)\}_{i=1}^{n}$. This proves (A).

Let $\bPsi^{(l)}$ be the $l$-th row of $\bPsi$, $l=1,...,m$. Denote $\bDelta=[\bdelta_1:\cdots:\bdelta_n]\in\mathcal{R}^{p\times n}$ and assume $\bz_i$ is the
$i$-th row of $\bDelta$. Using Lemma 2.9.5 in \cite{vanweak}, we obtain
\begin{align*}
\sqrt{p}\sum_{j=1}^{p}\Psi_{lj}\bz_j\rightarrow N(\bzero, \mbox{Cov}(\bz_1)).
\end{align*}
Therefore, $\sum_{j=1}^{p}\Psi_{lj}\bz_j=O_p(p^{-1/2})$, reducing the magnitude of noise in the original features. Hence (B) is proved. Thus, even if noise exists, asymptotic performance of $\{T(\bx_i)\}_{i=1}^n$ will be similar to $\{T(\bz_i)\}_{i=1}^n$ in the GP regression (which by (A) has ``optimal" asymptotic performance).

\section{Simulation Examples}

We assess the performance of compressed Gaussian process (CGP) regression in a number of simulation examples. We consider various numbers of features ($p$) and level of noise in the features ($\tau$) to study their impact on the performance. In all the simulations out of sample predictive performance of the proposed CGP regression was compared to that of uncompressed Gaussian process (GP), BART (Bayesian Additive Regression Trees) \cite{chipman2010bart}, RF (Random Forests) \cite{breiman2001random} and TGP (Treed Gaussian process) \cite{gramacy2008bayesian}. Unfortunately, with massive number of features, traditional BART, RF and TGP are computationally prohibitive.
Therefore, we consider compressed versions in which we generate a single projection matrix to obtain a single set of compressed features, running the analysis with compressed
features instead of original features. This idea leads to compressed versions of random forest (CRF), Bayesian additive regression tree (CBART) and Treed Gaussian process (CTGP). These methods entail faster implementation when the number of features is massive.

As a default in these analyses, we use $m=60$, which seems to be a reasonable choice of upper bound for the dimension of the linear subspace to compress to.
In addition, we implement two stage GP (2GP) where the $p$-dimensional features are projected into smaller dimension by using Laplacian eigenmap (\cite{belkin2003laplacian,guerrero2011laplacian}) in the first stage and then a GP with projected features is fitted in the second stage.
We also compared Lasso and partial least square regression (PLSR)  to indicate advantages of our proposed method over linear regularizing methods. However, in presence of strong nonlinear relationship between the response and the features, Lasso and PLSR perform poorly and hence results for them are omitted.

When $n$ is massive, to bypass heavy computational price associated with CGP for inverting an $n\times n$ matrix, we employ a low rank approximation of the compressed Gaussian process as described in section~\ref{sec:largen}. As an uncompressed competitor of CGP in settings with large $n$, efficient Gaussian random projection technique \cite{banerjee2013efficient} is implemented. This is also referred to as the GP to avoid needless confusion. Along with GP, CBART and CRF are included as competitors. CTGP with large $n$ poses heavy computational burden and is, therefore, omitted.

 As a more scalable competitor, we employ the popular two stage technique of clustering the massive sample into a number of clusters followed by fitting simple model such as Lasso in each of these clusters. To facilitate clustering of high dimensional features in the first stage, we use the spectral clustering algorithm (\cite{ng2001spectral}) described in Algorithm~\ref{alg1} .
\begin{algorithm*}[!ht]
{
\caption{Spectral Clustering Algorithm}
\label{alg1}
{\bf Input:} features $\bx_1,....,\bx_n$ and the number of clusters required $n.clust$.
\begin{itemize}
    \setlength{\itemsep}{0pt}%
    \setlength{\parskip}{0pt}
\item Form the affinity matrix $\bA\in\mathcal{R}^{n\times n}$ defined by $A_{ij}=\exp\left(-||\bx_i-\bx_j||^2/2\sigma^2\right)$ if $i\neq j$, $A_{ii}=0$, for some judicious choice of $\sigma^2$.
\item Define $\bD$ to be the diagonal matrix whose $(i,i)$-th entry is the sum of the elements in the $i$-th row of $\bA$. Construct $\bL=\bD^{-1/2}\bA\bD^{-1/2}$.
\item Find $\bs_1,...,\bs_{n.clust}$ be the eigenvectors corresponding to the $n.clust$ largest eigenvalues of $\bL$. Form the matrix
$\bS=[\bs_1:\cdots:\bs_{n.clust}]\in \mathcal{R}^{n\times n.clust}$ by stacking the eigenvectors in column.
\item Normalize so that each row of $\bS$ has unit norm.
\item Now treating each row of $\bS$ as a point in $\mathcal{R}^{n.clust}$ cluster them into $n.clust$ clusters via $K-means$ clustering.
\item Finally assign $\bx_i$ in cluster $j$ if the $i$ th row of $\bS$ goes to cluster $j$.
\end{itemize}
}
\end{algorithm*}
Once observations are clustered, separate Lasso is fitted in each of these clusters. Henceforth, we refer to this procedure as distributed supervised learning (DSL).

The model averaging step in CGP requires choosing a window over the possible values of $m$. When $n$ is small, we adopt the choice suggested in \cite{guhaniyogi2013bayesian} to have a window of $[\ceil{2log(p)},min(n,p)]$, which implies that the number of possible models to be averaged across is $s=min(n,p)-\ceil{2log(p)}+1$. When $n$ is large, we choose the window of $[\ceil{2log(p)},min(m_{\bPhi},p)]$. The number of rows of $\bPhi$ is fixed at $m_{\bPhi}=150$ for the simulation study with large $n$. However, changing $m_{\bPhi}$ moderately does not alter the performance of CGP.

\subsection{Manifold Regression on the Swiss Roll}\label{sec:smallsim}

\begin{figure}[h]
\centering
\subfigure[noise corrupted swiss roll]{\label{swissa}\includegraphics[width=0.4\columnwidth]{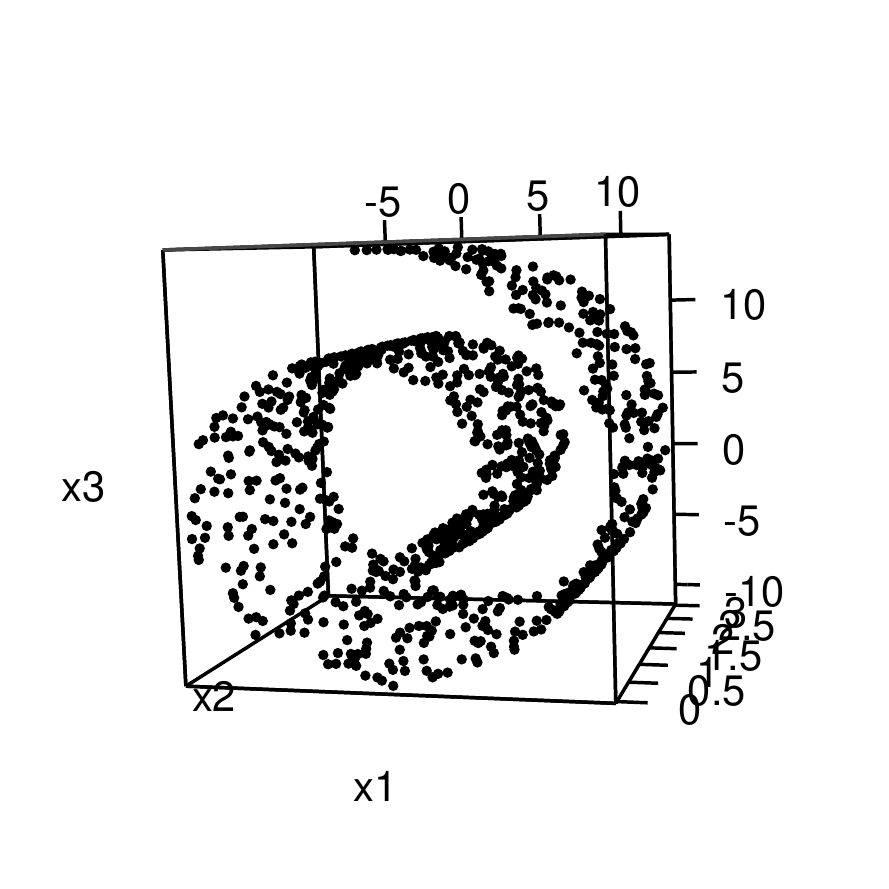}}
\subfigure[response vs. $x_1,x_2$]{\label{swissb}\includegraphics[width=0.4\columnwidth]{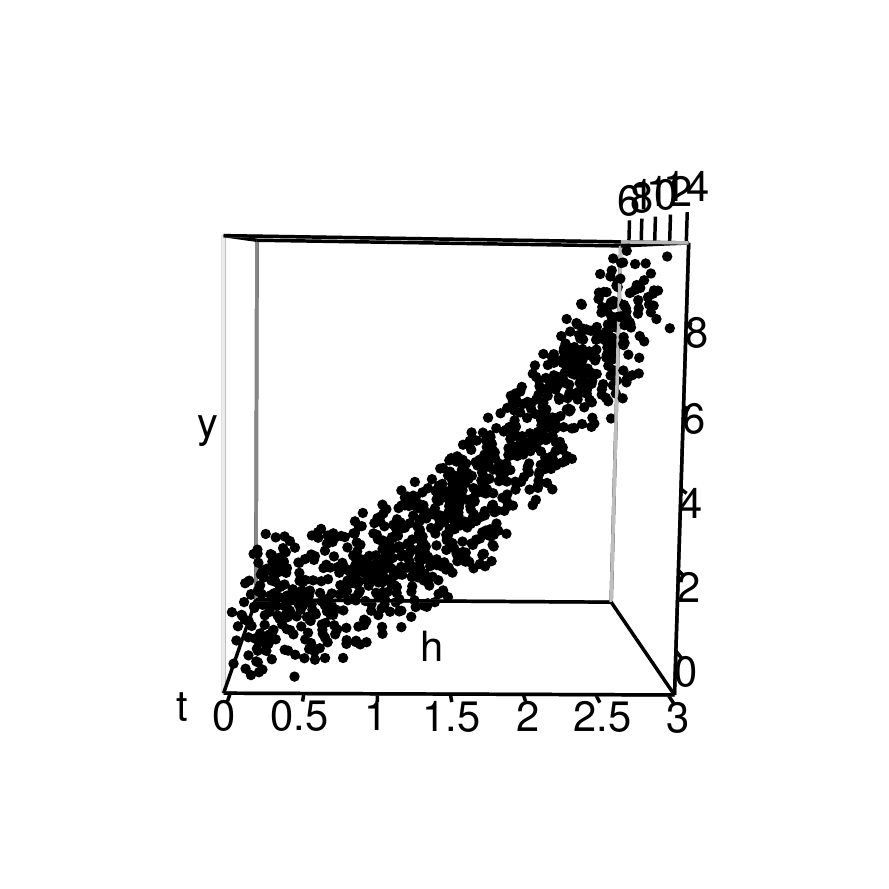}}\\
\subfigure[swiss roll shown in 2d]{\label{swissc}\includegraphics[width=0.4\columnwidth]{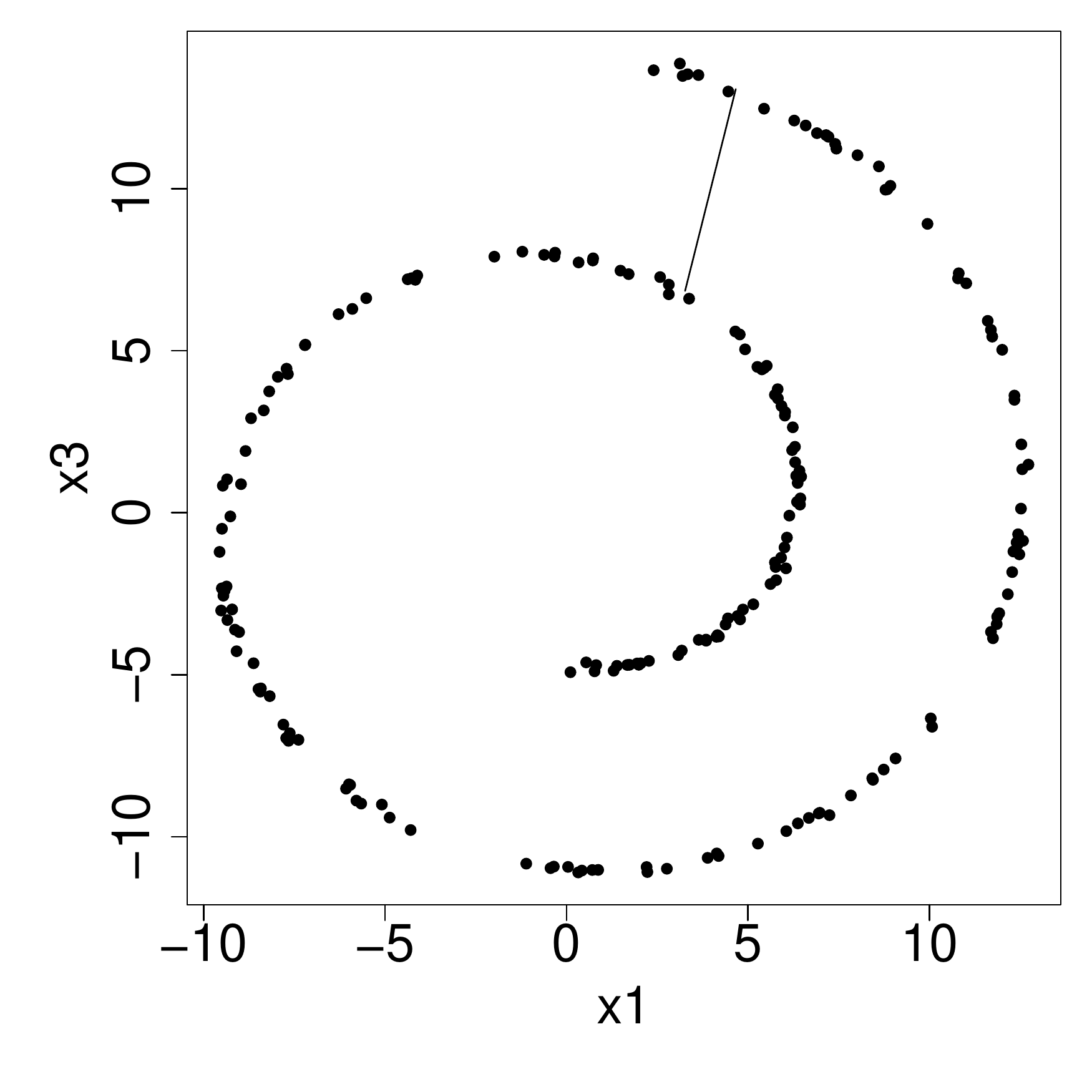}}
\caption{Simulated features and response on a \emph{noisy} Swiss Roll, $\tau=0.05$ }\label{fig11}
\end{figure}

To provide some intuition for our model, we start with a concrete example where the distribution of the response is a nonlinear function of
the coordinates along a swissroll, which is embedded in a high dimensional ambient space. To be more specific, we sample manifold coordinates,
$t\sim U(\frac{3\pi}{2},\frac{9\pi}{2})$, $h\sim U(0,3)$. A high dimensional feature $\bx=(x_1,...,x_p)$ is then sampled
following
\begin{align*}
x_1 =t\cos(t)+\delta_1,\:x_2=h+\delta_2,\:x_3=t\sin(t)+\delta_3,\:x_i=\delta_i,\:i\geq 4,
\delta_1,..,\delta_p \sim N(0,\tau^2).
\end{align*}
Finally responses are simulated to have nonlinear and non-monotonic relationship with the features
\begin{align}\label{eq:respsim}
y_i=\sin(5\pi t)+h^2+\epsilon_i,\:\epsilon_i\sim N(0,0.02^2).
\end{align}
Clearly, $\bx$ and $y$ are conditionally independent given $\theta, h$, which is the low-dimensional signal manifold. In particular,
$\bx$ lives on a (\emph{noise corrupted}) swissroll embedded in a $p$-dimensional ambient space (see Figure~\ref{swissa}), but $y$
is only a function of coordinates along the swissroll $\mathcal{M}$ (see Figure~\ref{swissb}).

The geodesic distance between two points in a swiss roll can be substantially different from their Euclidean distance in the ambient space $\mathcal{R}^p$. For example, in Figure~\ref{swissc} two points joined by the line segment have much smaller Euclidean distance than geodesic distance. Theorem~\ref{th:diffeomorph} in Section~\ref{sec:convergencerate} guarantees optimal performance when the compact sub-manifold $\mathcal{M}$ is sufficiently smooth, so that the locally Euclidean distance serves as a good approximation of the geodesic distance. The Swiss roll presents a challenging set up for CGP, since points on $\mathcal{M}$ that are close in a Euclidean sense can be quite far in a geodesic sense.

To assess the impact of the number of features (p) and noise levels of the features ($\tau$) on the performance of CGP, a number of simulation scenarios are considered in Table~\ref{tab:sim_runs}. For each of these simulation scenarios, we generate multiple datasets and present predictive inference such as mean squared prediction error (MSPE), coverage and lengths of 95\% predictive intervals (PI) averaged over all replicates.
{\footnotesize
\begin{table}[h]
\centering
\begin{tabular}{cccc}
	\hline
Simulation & sample size ($n$) & no. of features ($p$) & noise in the features ($\tau$)\\
	\hline
1 & 100  & 10,000 & 0.02\\
2 & 100  & 20,000 & 0.02\\
3 & 100  & 10,000 & 0.05\\
4 & 100  & 20,000 & 0.05\\
5 & 100  & 10,000 & 0.10\\
6 & 100  & 20,000 & 0.10\\
	\hline
\end{tabular}
\caption{Different Simulation settings for CGP.}
\label{tab:sim_runs}
\end{table}}

In our experiments, $\by$ and $\bX$ are centered.
To implement LASSO, we use \texttt{glmnet} (\cite{friedman2009glmnet}) package in R with the optimal tuning parameter selected through 10 fold cross validation.
 CRF, CBART and CTGP in R using \texttt{randomForest} (\cite{liaw2002classification}), \texttt{BayesTree}
(\cite{chipman2009package}) and \texttt{tgp} (\cite{gramacy2007tgp}) packages, respectively.

\subsubsection{MSPE Results}
Predictive MSE for each of the simulation settings averaged over 10 simulated datasets is
shown in Table~\ref{tab2}. Subscripted values represent bootstrap standard errors for the averaged MSPEs,
calculated by generating 50 bootstrap datasets resampled from the MSPE values, finding the
average MSPE of each, and then computing their standard error.

Table~\ref{tab2} shows that feeding randomly compressed features into any of the nonparametric methods leads to good predictive performance, while Lasso fails to improve much upon the null model (not shown here). For both $p=10,000, 20,000$, when the swiss roll is corrupted with low noise, CGP, CBART and CRF provide significantly better performance than GP. Increasing noise in the features results in deteriorating performances for all the competitors. CGP is an effective tool to reduce the effect of noise in the features, but at a {\em tipping point}  (depending on $n$) noise distorts the manifold too much,
and CGP starts performing similarly to GP.  CRF and CTGP perform much worse than CGP in high noise scenarios, while CBART produces competitive performance.
Two stage GP (2GP) performs much worse than all the other competitors; perhaps the two stage procedure is considerably more sensitive to noise.
Increasing number of features does not alter MSPE for CGP significantly in presence of low noise, consistent with asymptotic results showing posterior convergence rates depend on the intrinsic dimension of $\mathcal{M}$ instead of $p$ when features are concentrated close to $\mathcal{M}$.  In the next section, we will study these aspects with increasing sample size and noise in the features.

\begin{table}[h]
\centering
\begin{tabular}{|  l | l | c | c | c | }
\cline{1-5}
 & \multicolumn{4}{ |c| }{Noise in the feature} \\
\cline{2-5}
 &  & .02 & .05 & .10 \\ \hline
\multirow{6}{*}{$p=10000$} & CGP &  $4.85_{0.24}$ & $5.76_{0.19}$ & $6.81_{0.24}$ \\ 
 & GP & $5.24_{0.23}$ & $5.62_{0.20}$ & $6.65_{0.25}$ \\
 & CRF & $4.72_{0.31}$ & $6.15_{0.23}$ & $7.12_{0.24}$ \\
 & CBART & $4.16_{0.41}$ & $6.18_{0.17}$ & $6.99_{0.22}$ \\
 & CTGP & $4.87_{0.47}$ & $7.25_{0.13}$ & $7.22_{0.20}$ \\
 & 2GP  & $6.26_{0.64}$ & $7.14_{0.29}$ & $7.65_{0.22}$    \\
 \hline
\multirow{6}{*}{$p=20000$} & CGP &  $5.02_{0.19}$ & $6.61_{0.24}$ & $7.58_{0.20}$ \\ 
 & GP & $5.29_{0.21}$ & $6.44_{0.23}$ & $7.46_{0.27}$ \\
 & CRF & $4.90_{0.26}$ & $6.75_{0.25}$ & $7.68_{0.30}$ \\
 & CBART & $4.56_{0.29}$ & $6.80_{0.19}$ & $7.61_{0.24}$ \\
 & CTGP & $6.18_{0.32}$ & $7.41_{0.23}$ & $7.84_{0.22}$ \\
 & 2GP  & $6.32_{0.52}$ & $6.99_{0.45}$  & $7.06_{0.38}$ \\
\hline
\end{tabular}
\caption{Performance comparisons for competitors in terms of mean squared prediction errors (MSPE)}\label{tab2}
\end{table}


\subsubsection{Coverage and Length of PIs}

\begin{figure}[h]
\centering
    \subfigure[$n=100$, $\tau=.02$, $p=10,000$]{\label{coverDt}\includegraphics[width=7cm]{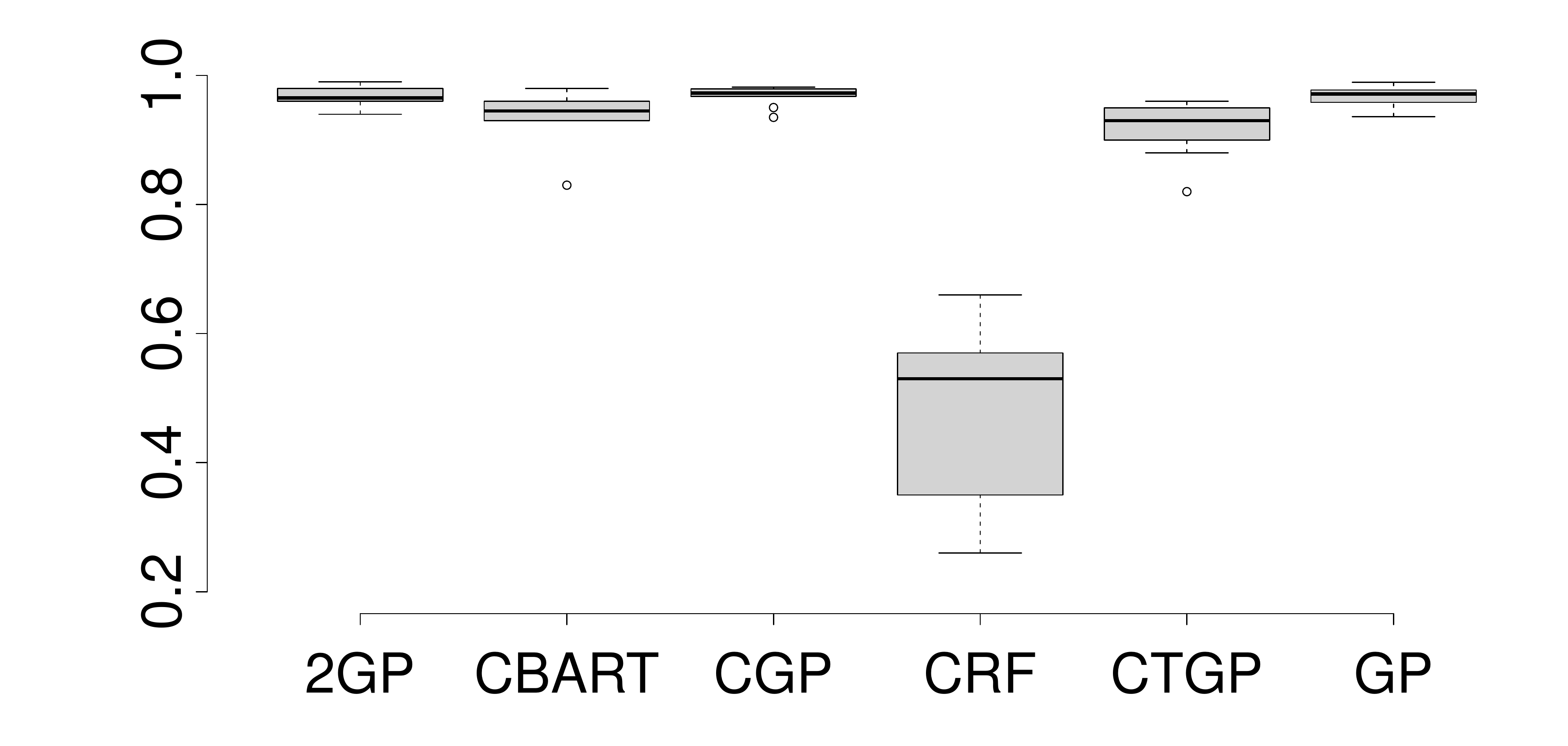}}
    \subfigure[$n=100$, $\tau=.02$, $p=20,000$]{\label{coverEt}\includegraphics[width=7cm]{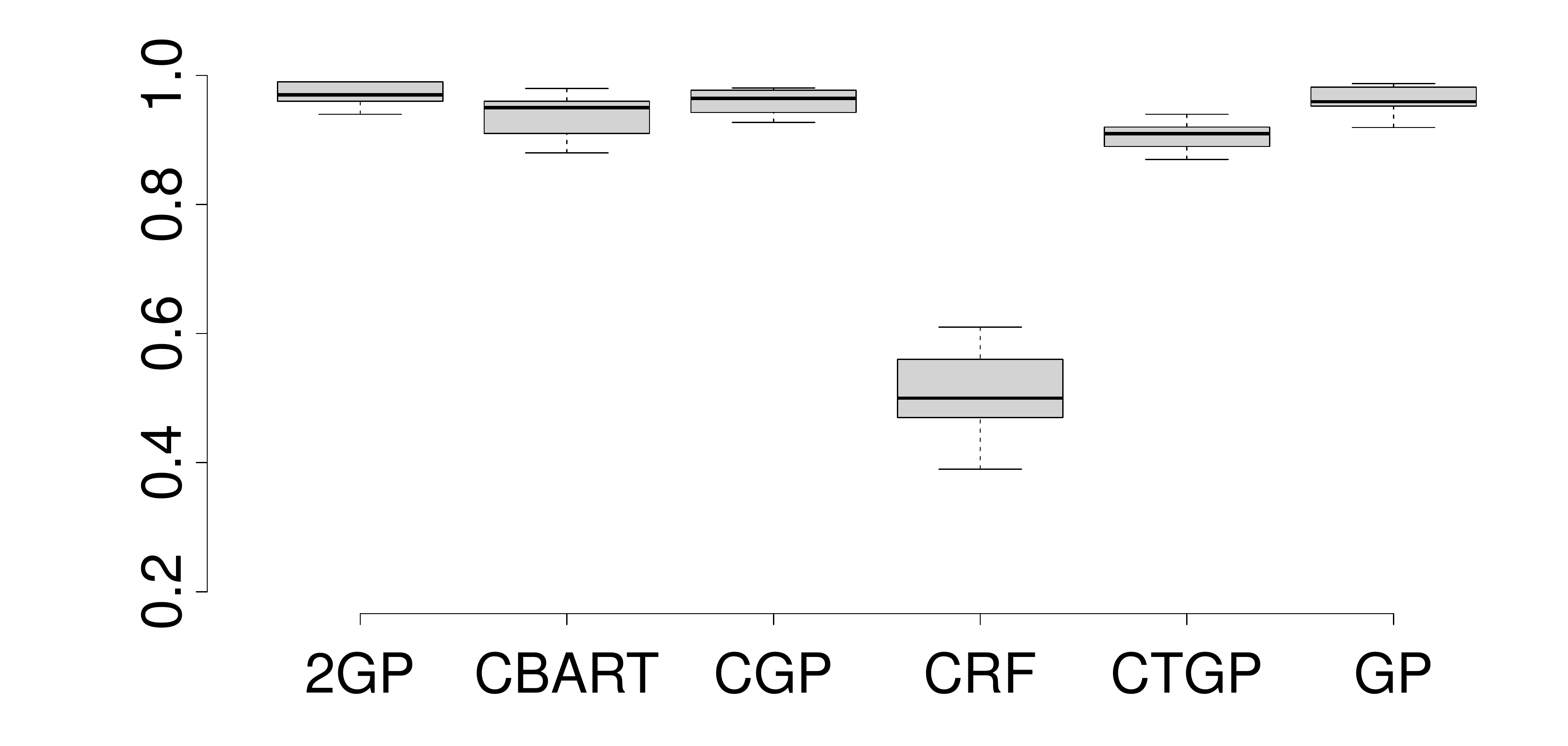}}\\
    \subfigure[$n=100$, $\tau=.05$, $p=10,000$]{\label{coverFt}\includegraphics[width=7cm]{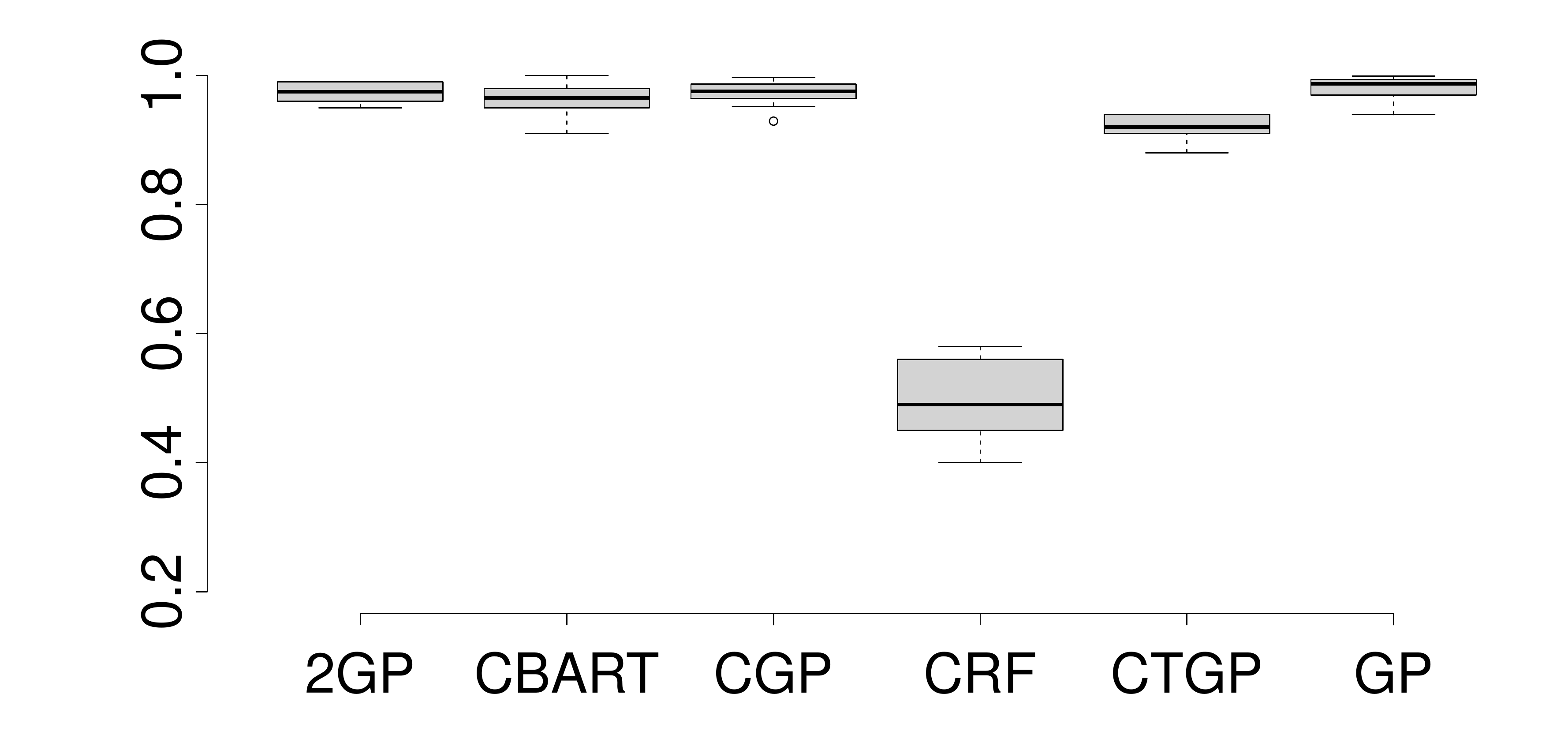}}
    \subfigure[$n=100$, $\tau=.05$, $p=20,000$]{\label{coverGt}\includegraphics[width=7cm]{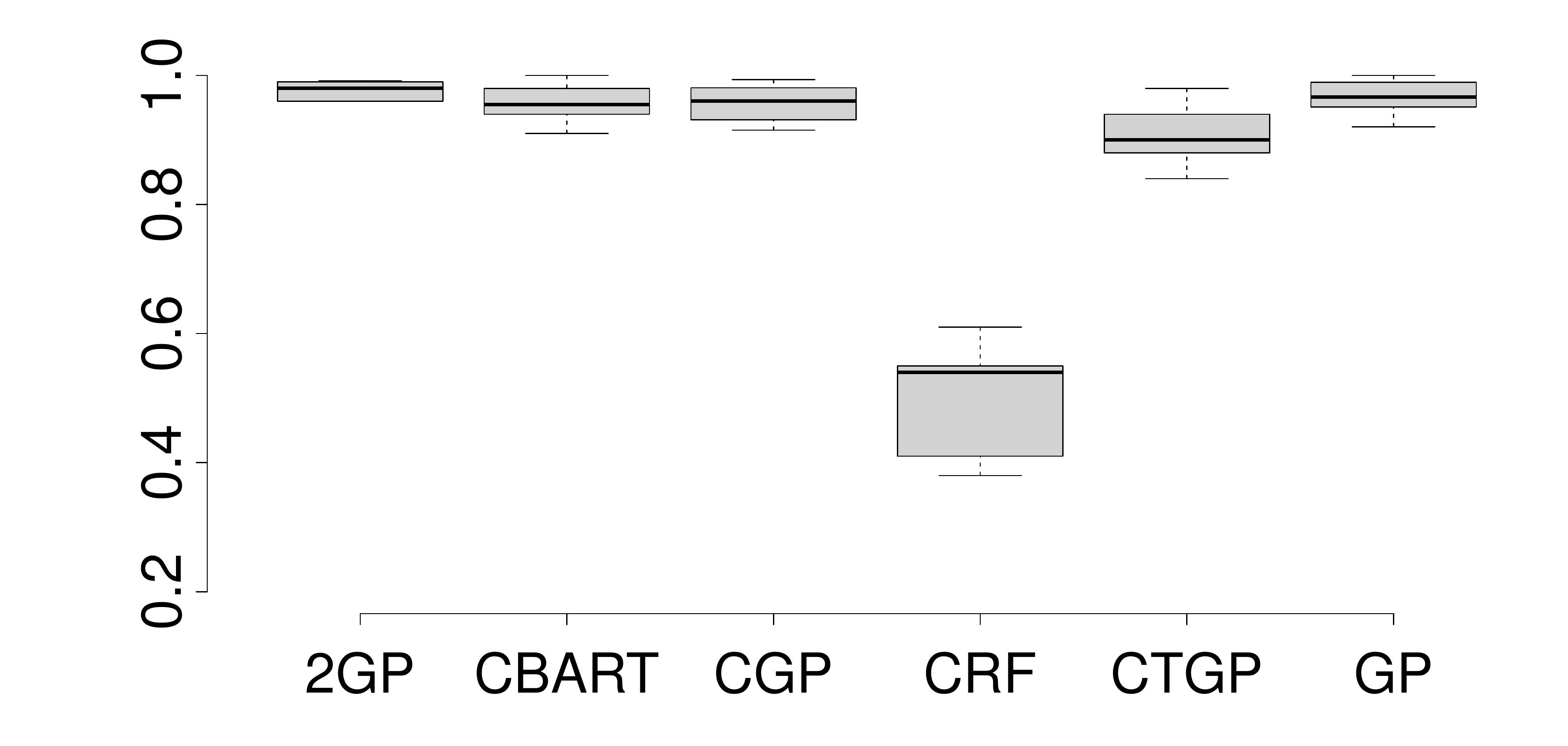}}\\
    \subfigure[$n=100$, $\tau=.10$, $p=10,000$]{\label{coverDt}\includegraphics[width=7cm]{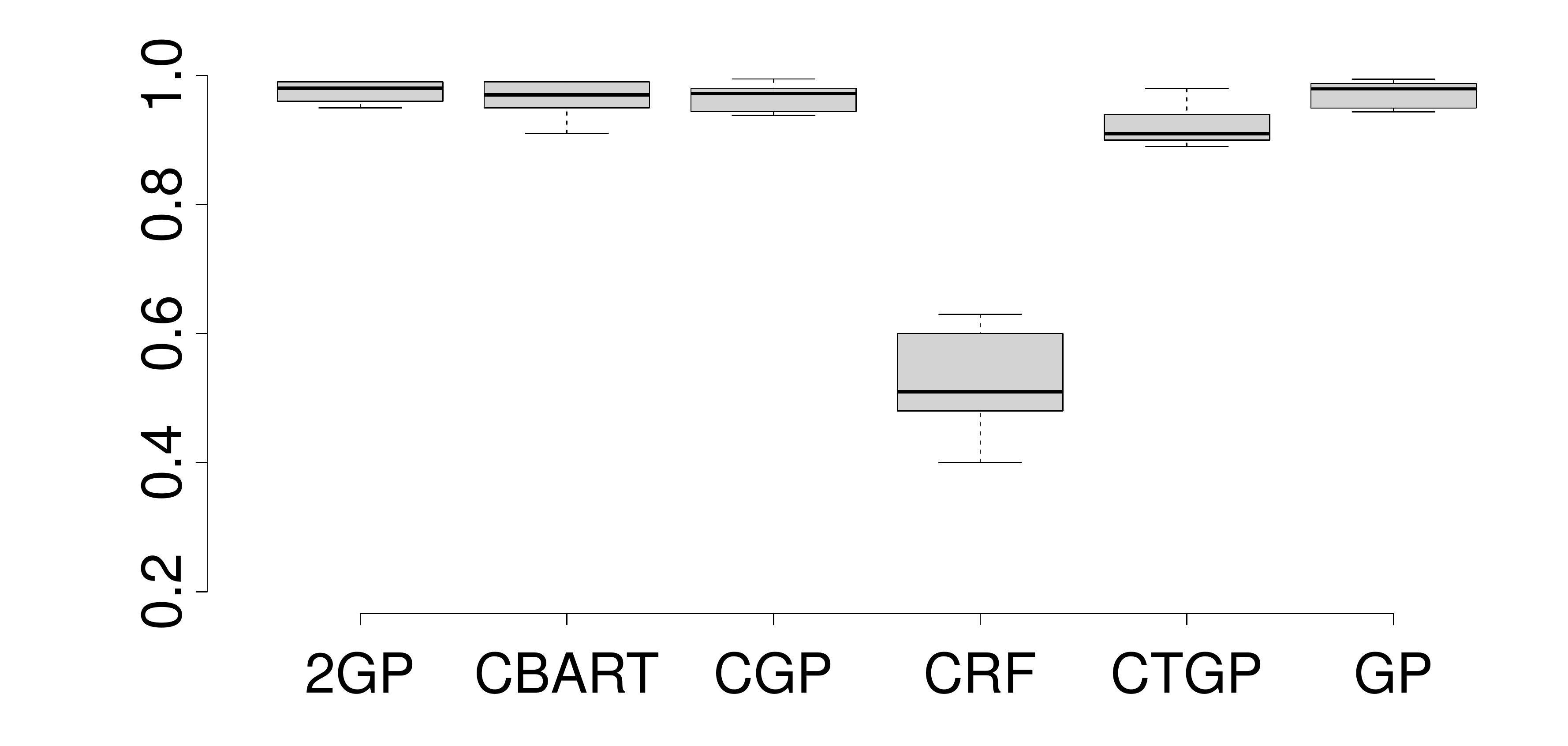}}
    \subfigure[$n=100$, $\tau=.10$, $p=20,000$]{\label{coverEt}\includegraphics[width=7cm]{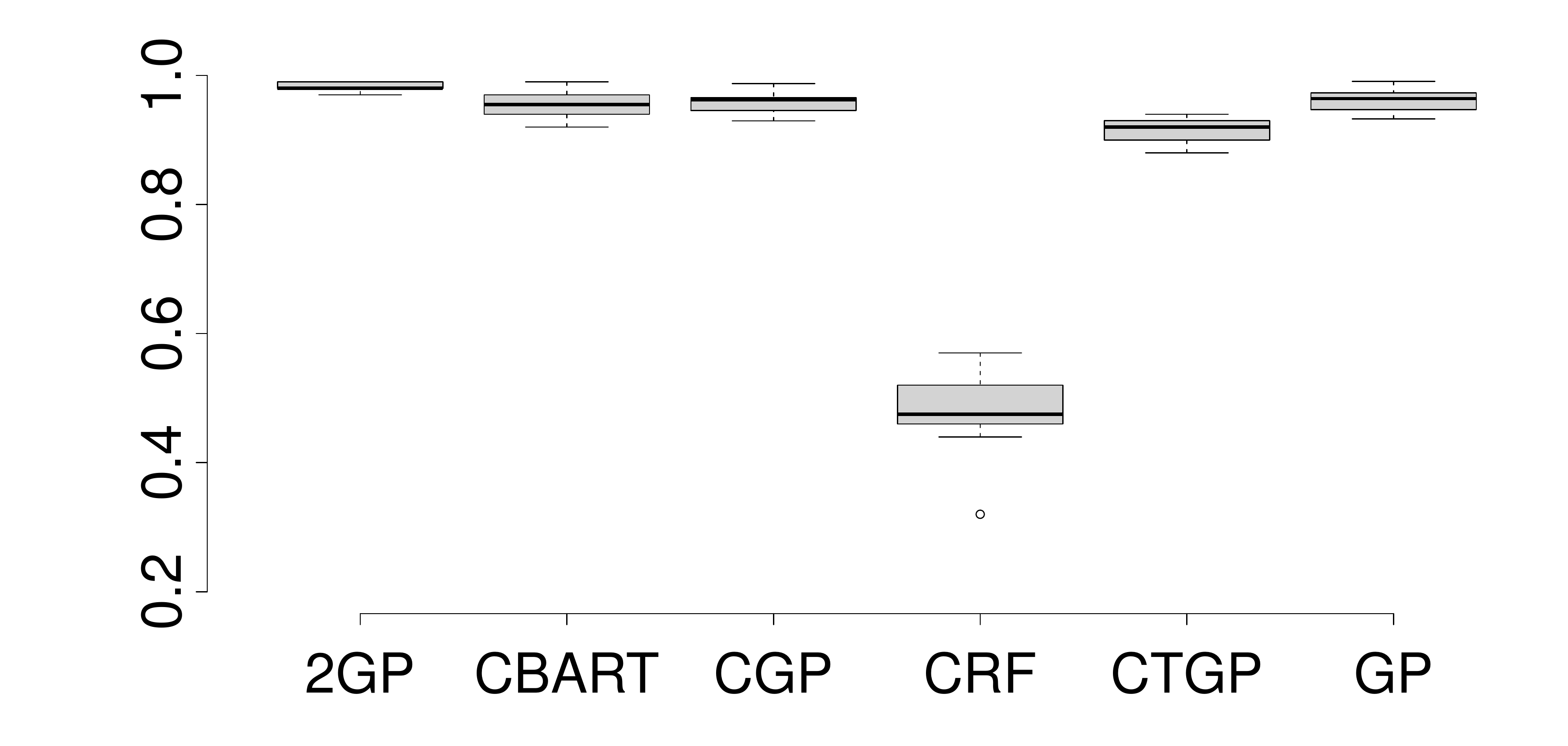}}
\caption{coverage of 95\% PI's for CGP, GP, CBART, CTGP, CRF, 2GP}\label{fig10}
\end{figure}

\begin{figure}[h]
\centering
    \subfigure[$n=100$, $\tau=.02$, $p=10,000$]{\label{lengDt}\includegraphics[width=7cm]{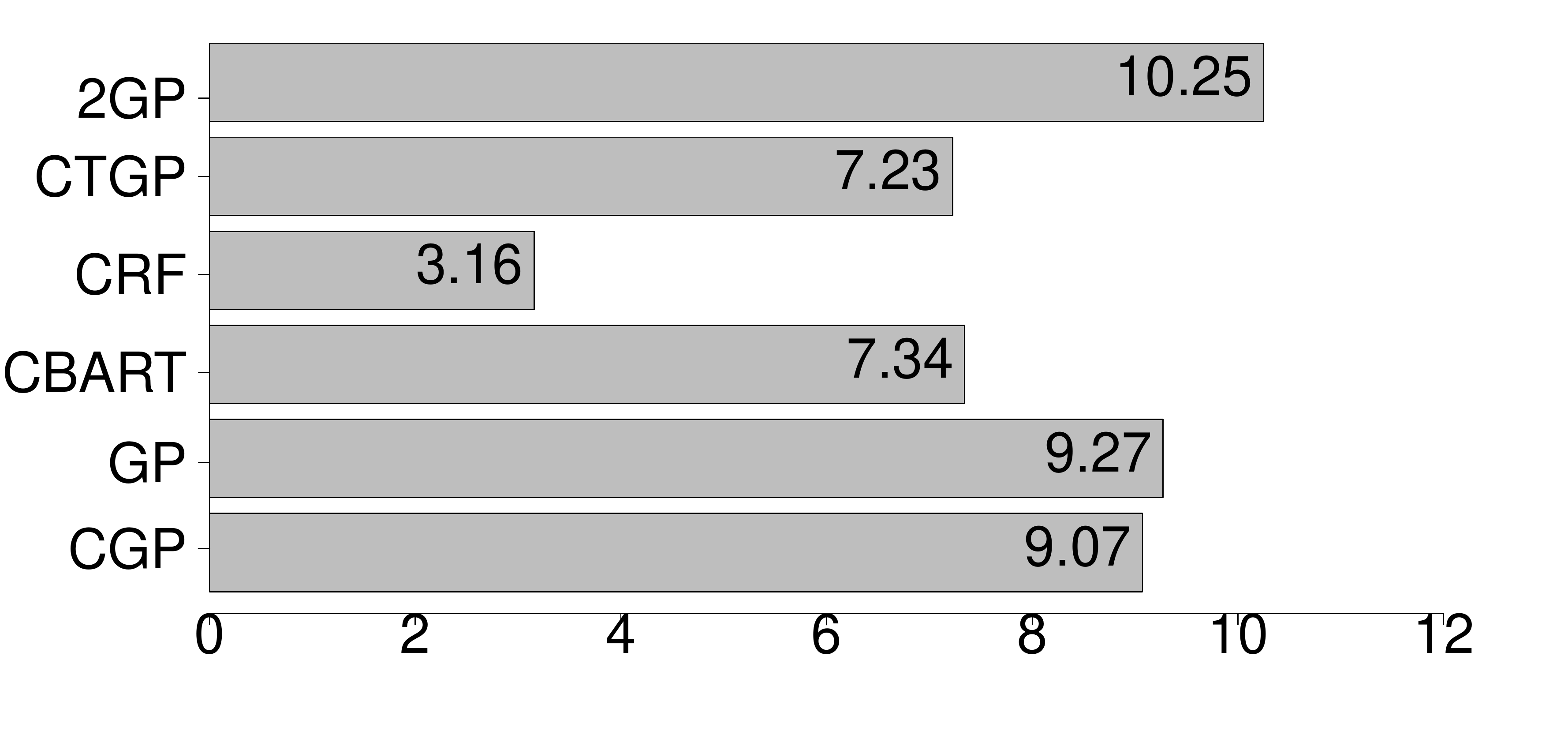}}
    \subfigure[$n=100$, $\tau=.02$, $p=20,000$]{\label{lengEt}\includegraphics[width=7cm]{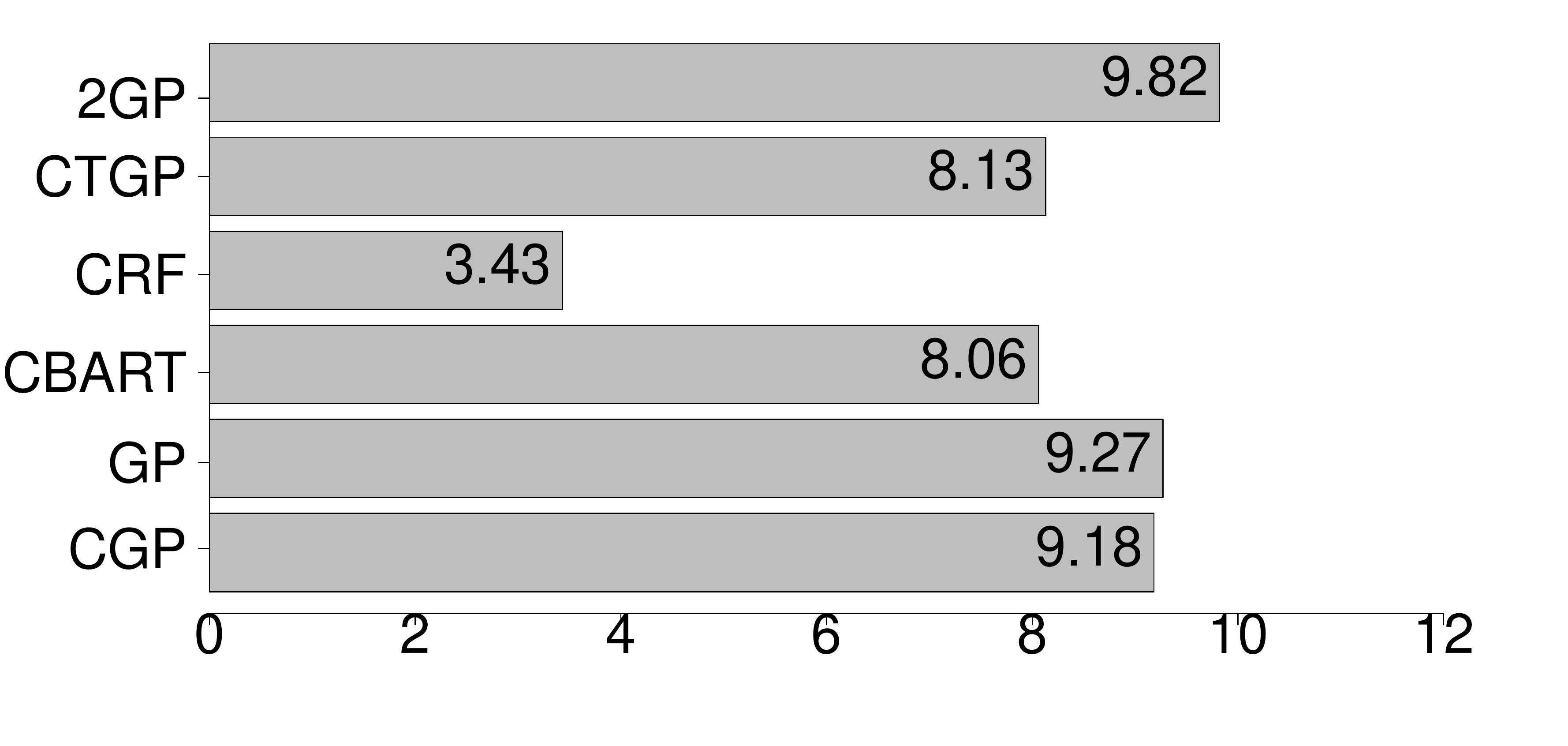}}\\
    \subfigure[$n=100$, $\tau=.05$, $p=10,000$]{\label{lengFt}\includegraphics[width=7cm]{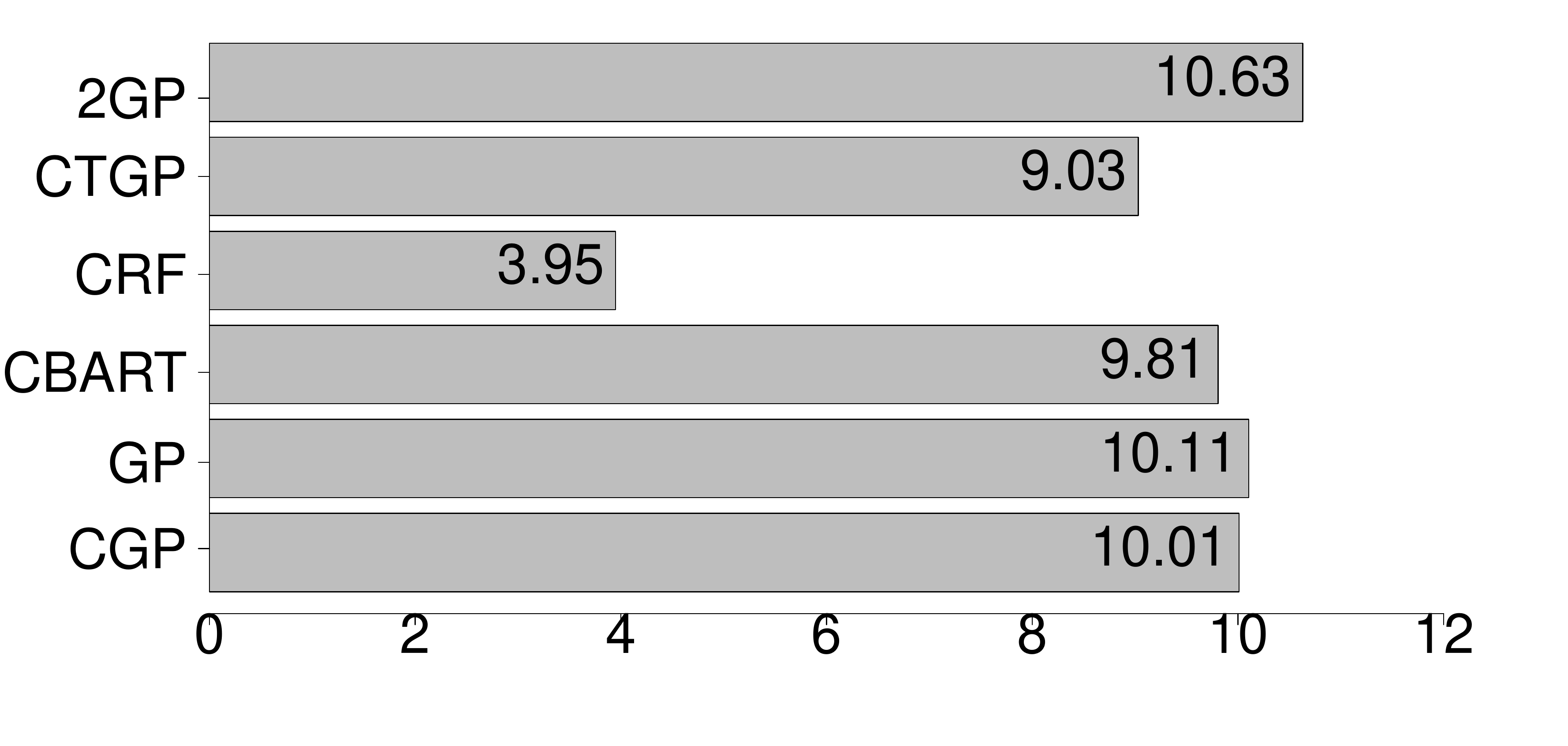}}
    \subfigure[$n=100$, $\tau=.05$, $p=20,000$]{\label{lengGt}\includegraphics[width=7cm]{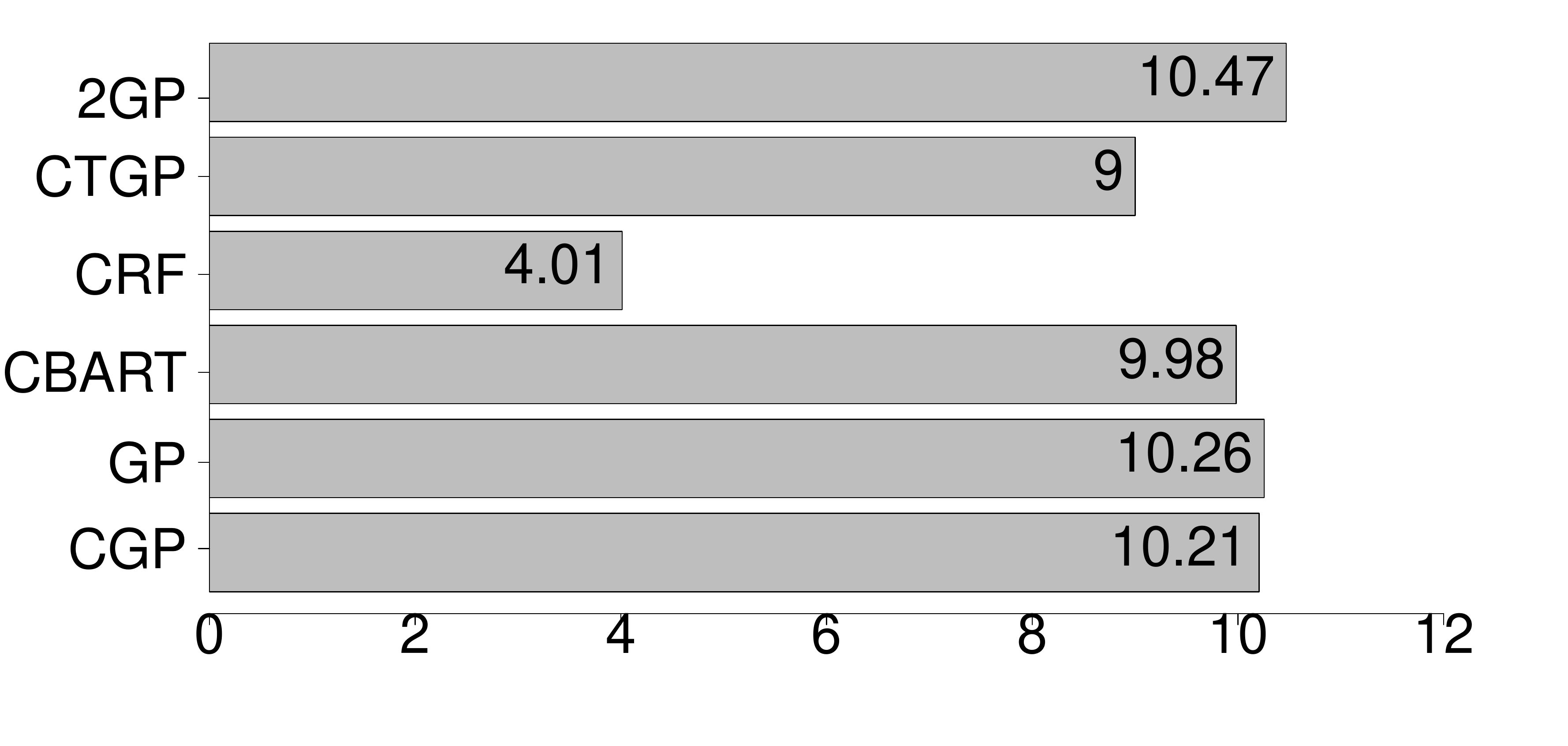}}\\
    \subfigure[$n=100$, $\tau=.10$, $p=10,000$]{\label{lengDt}\includegraphics[width=7cm]{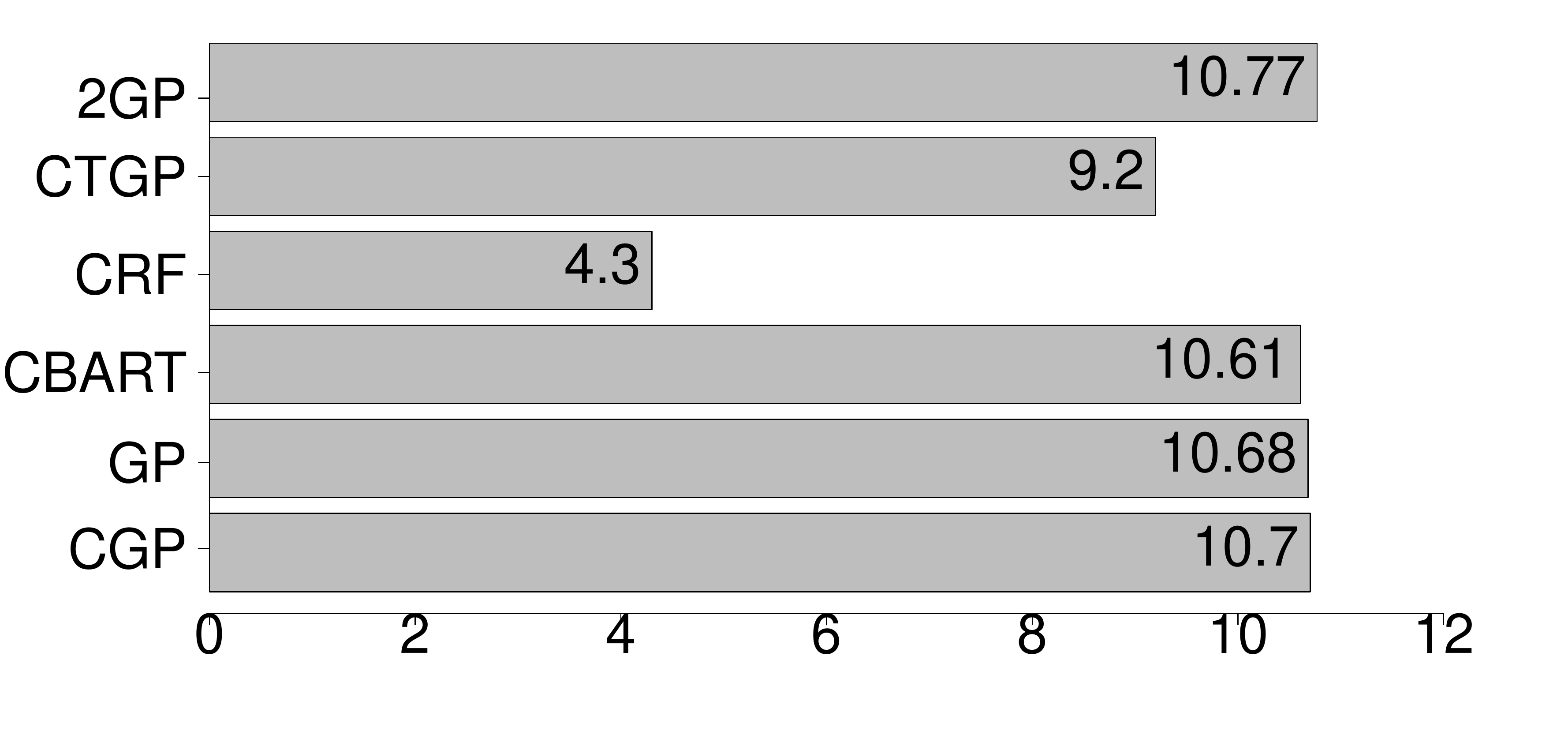}}
    \subfigure[$n=100$, $\tau=.10$, $p=20,000$]{\label{lengEt}\includegraphics[width=7cm]{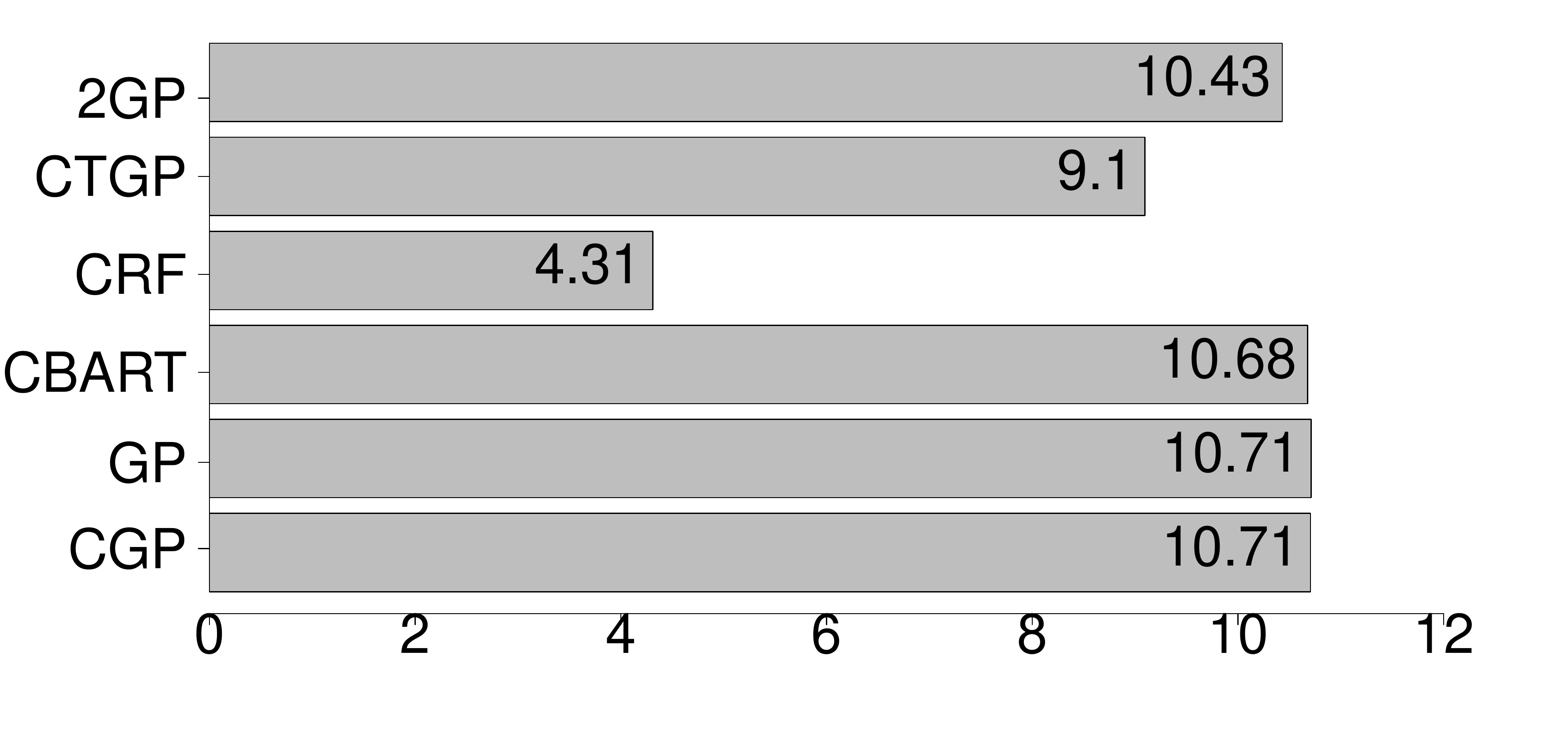}}
\caption{lengths of 95\% PI's for CGP, GP, CBART, CRF, CBART, 2GP}\label{fig15}
\end{figure}

To assess if CGP is well calibrated in terms of uncertainty quantification, we compute coverage and length of 95\% predictive intervals (PI) of CGP along with all the competitors. Although most frequentist methods such as CRF are unable to provide such coverage probabilities in producing point estimates, we present a measure of predictive uncertainty for those methods following the popular two stage plug-in approach, (i) estimate the regression function in the first stage; (ii) construct 95\% PI  based on the normal distribution centered on the predictive mean from the regression model with variance equal to the estimated variance in the residuals.  Boxplots for coverage probabilities in all the simulation cases are presented in Figure~\ref{fig10}. Figure~\ref{fig15} presents median lengths of the 95\% predictive intervals.

Both these figures demonstrate that in all the simulation scenarios CGP, uncompressed GP, 2GP and CBART
result in predictive coverage of around 95\%, while CRF suffers from severe under-coverage. The gross under-coverage of CRF is attributed to the overly narrow
predictive intervals. Additionally, CTGP shows some under-coverage, with shorter predictive intervals than CGP, GP, 2GP or CBART.
CGP turns out to be an excellent choice  among all the competitors in fairly broad simulation scenarios.  We consider larger sample sizes and high noise scenarios in the next subsection.

\subsection{Manifold Regression on Swiss roll for Large Sample}

To assess how the relative performance of CGP changes for larger sample size, we implement manifold regression on swiss roll using methodologies developed in section 3.
For this simulation example, data generation scheme similar to section~\ref{sec:smallsim} is used. Ideally, larger sample
size should lead to better predictive performance. Therefore, one would expect more accurate prediction even with higher degree of noise in the features for larger sample size, as long as there is sufficient signal in the data. To accommodate higher signal than in section~\ref{sec:smallsim}, we simulate manifold coordinates as $t\sim U(\frac{3\pi}{2},\frac{9\pi}{2})$, $h\sim U(0,5)$ and sample responses as per (\ref{eq:respsim}). We also increase noise variability in the features for all the simulation settings. Simulation scenarios are described in Table~\ref{tab2:sim_runs}.

{\footnotesize
\begin{table}[h]
\centering
\begin{tabular}{cccc}
	\hline
Simulation & sample size ($n$) & no. of features ($p$) & noise in the features ($\tau$)\\
	\hline
1 & 5,000  & 10,000 & .03\\
2 & 5,000  & 20,000 & .03\\
3 & 5,000  & 10,000 & .06\\
4 & 5,000  & 20,000 & .06\\
5 & 5,000  & 10,000 & .10\\
6 & 5,000  & 20,000 & .10\\
	\hline
\end{tabular}
\caption{Different Simulation settings for CGP for large $n$.}
\label{tab2:sim_runs}
\end{table}}
MSPE of all the competing methods are calculated along with their bootstrap standard errors and presented in Table~\ref{tab12}. Results in Table~\ref{tab12} provide more evidence supporting our conclusion in section \ref{sec:smallsim}. With smaller noise variance, CGP along with other compressed methods outperform uncompressed GP and 2GP. As $\tau$ increases, the manifold structure is more and more disrupted, with performance
of all the competitors worsening. With increasing noise variance, performance of CGP and GP start becoming comparable, while the other compressed methods provide inferior performance. Comparing results from the last section it is quite evident that with large samples, CGP is able to perform
well even with very large number of features and moderate variance of noise in the features. In large samples, the performance difference between GP and CGP becomes drastic with small noise variance. This shows the effectiveness of CGP for large $n,p$ when features are close to lying on a low-dimensional manifold.

\begin{figure}[h]
\centering
    \subfigure[$n=5,000$, $p=10,000$, $\tau=0.03$]{\label{coverDt}\includegraphics[width=7cm]{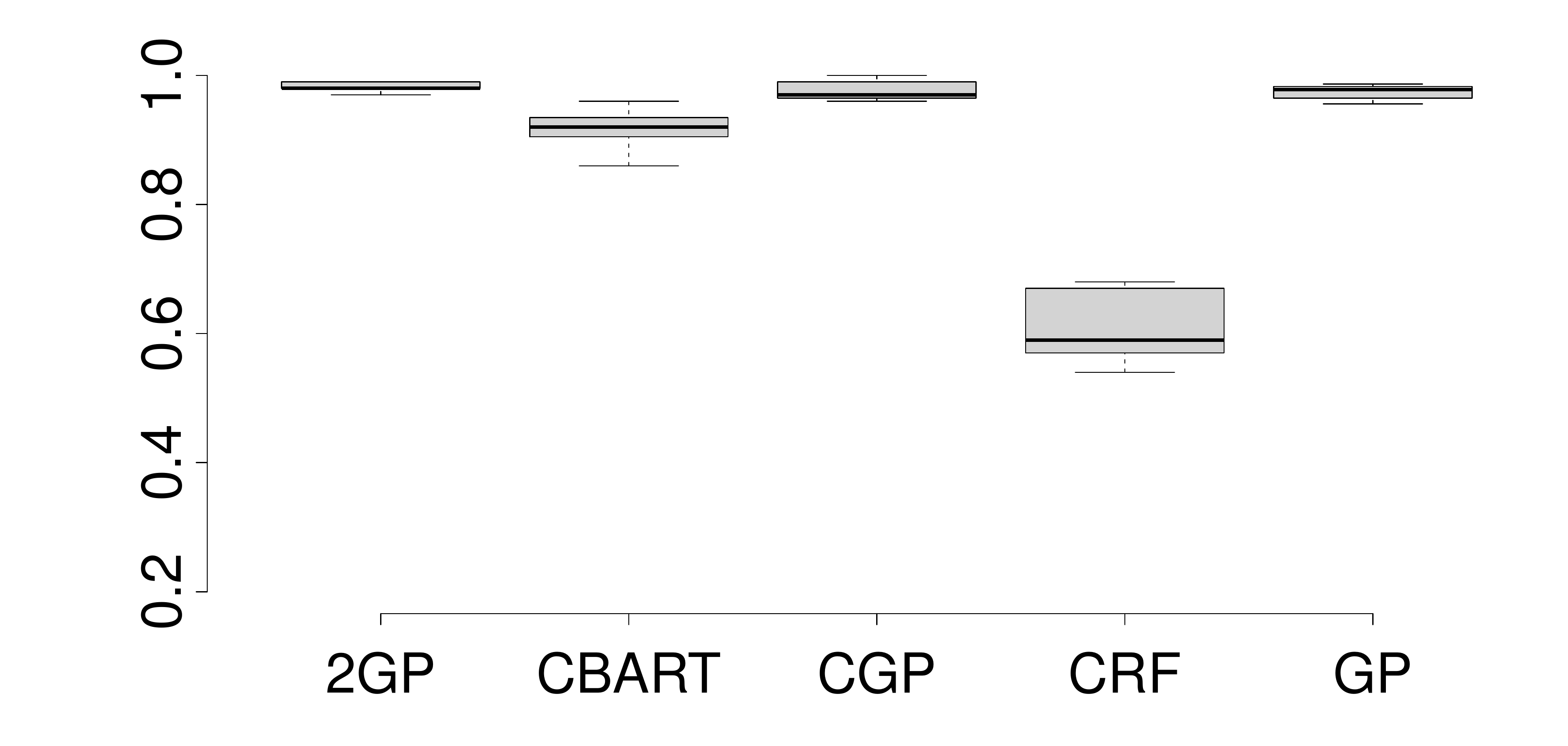}}
    \subfigure[$n=5,000$, $p=20,000$, $\tau=0.03$]{\label{coverEt}\includegraphics[width=7cm]{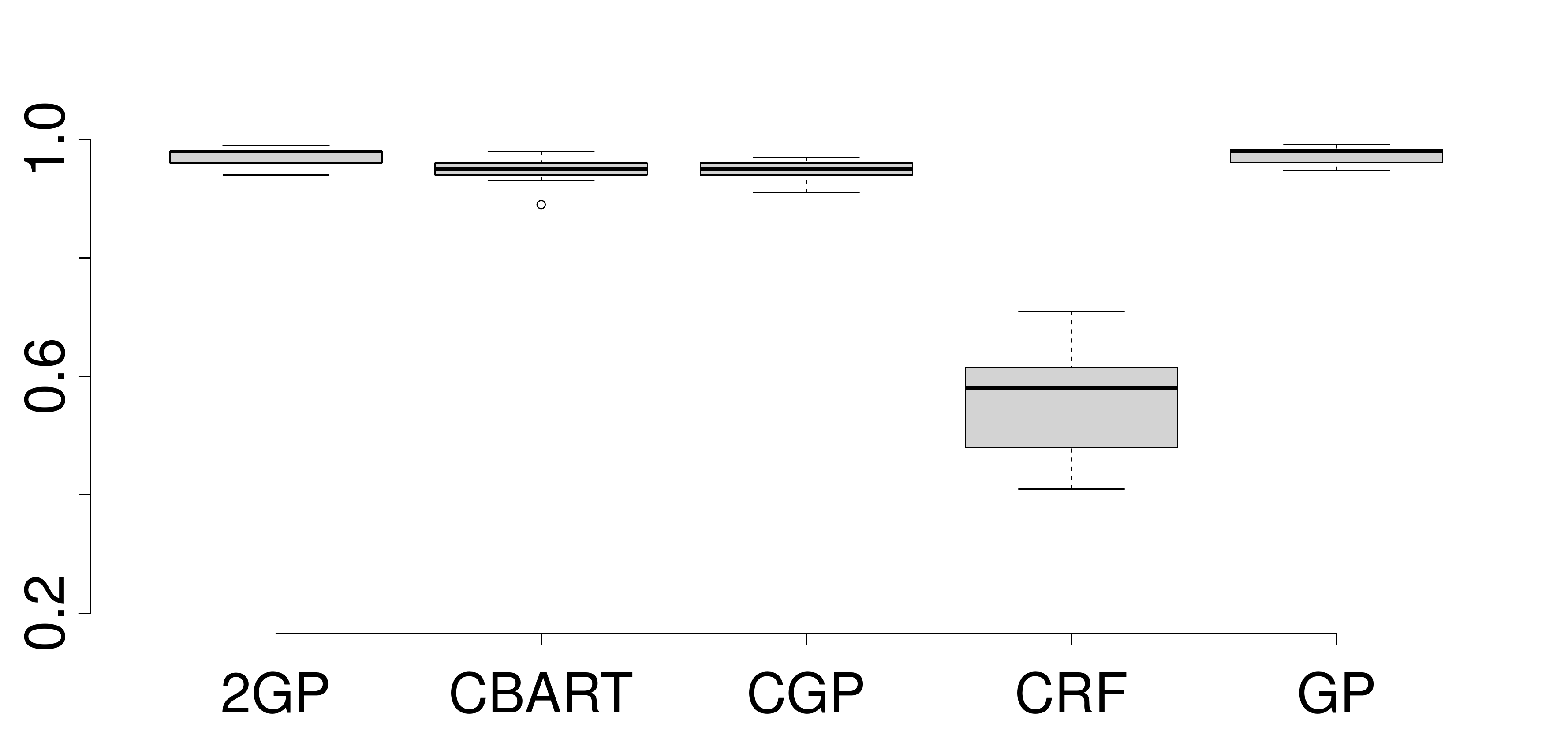}}\\
    \subfigure[$n=5,000$, $p=10,000$, $\tau=0.06$]{\label{coverGt}\includegraphics[width=7cm]{cov50001000003.pdf}}
    \subfigure[$n=5,000$, $p=20,000$, $\tau=0.06$]{\label{coverFt}\includegraphics[width=7cm]{cov50002000003.pdf}}\\
    \subfigure[$n=5,000$, $p=10,000$, $\tau=0.10$]{\label{coverHt}\includegraphics[width=7cm]{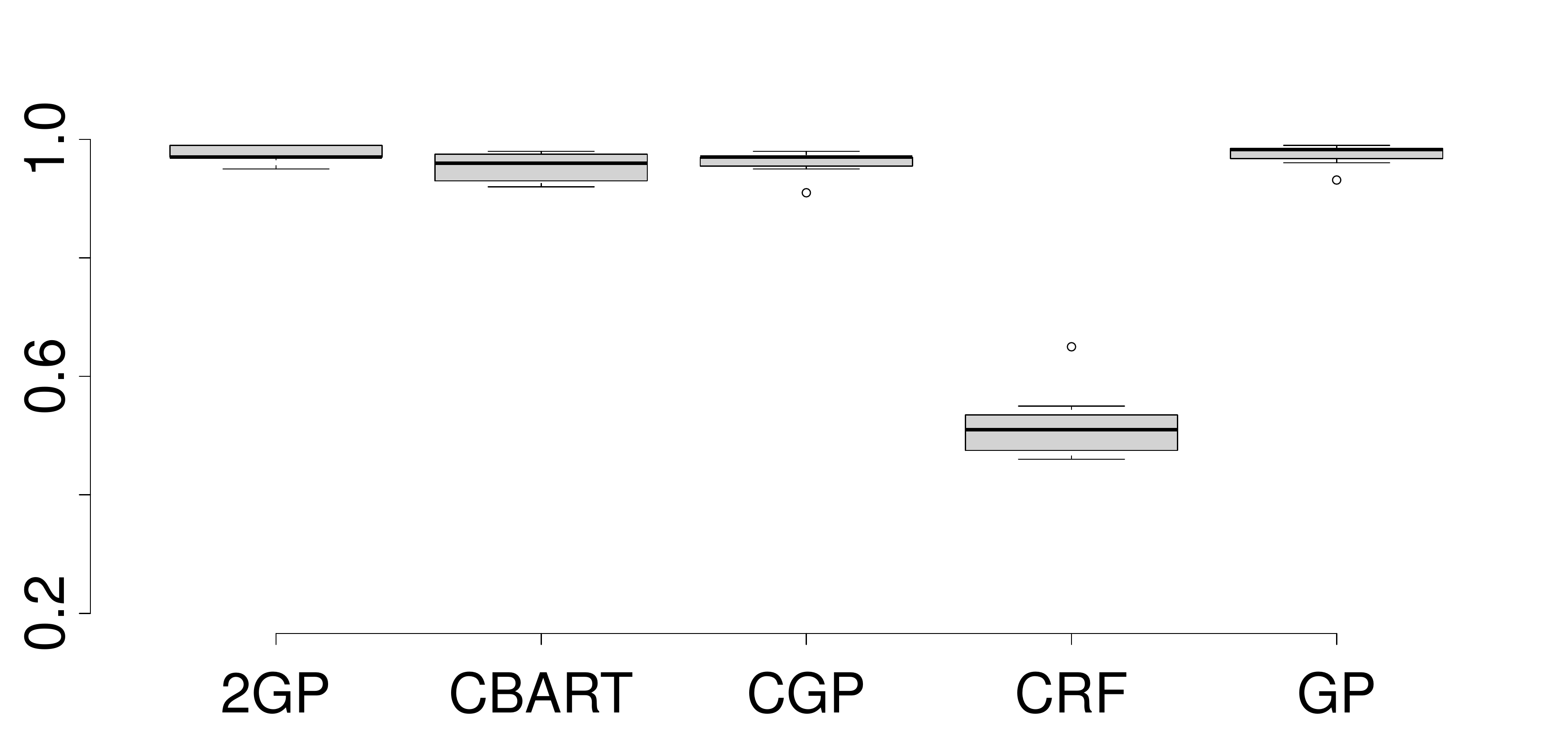}}
    \subfigure[$n=5,000$, $p=20,000$, $\tau=0.10$]{\label{coverIt}\includegraphics[width=7cm]{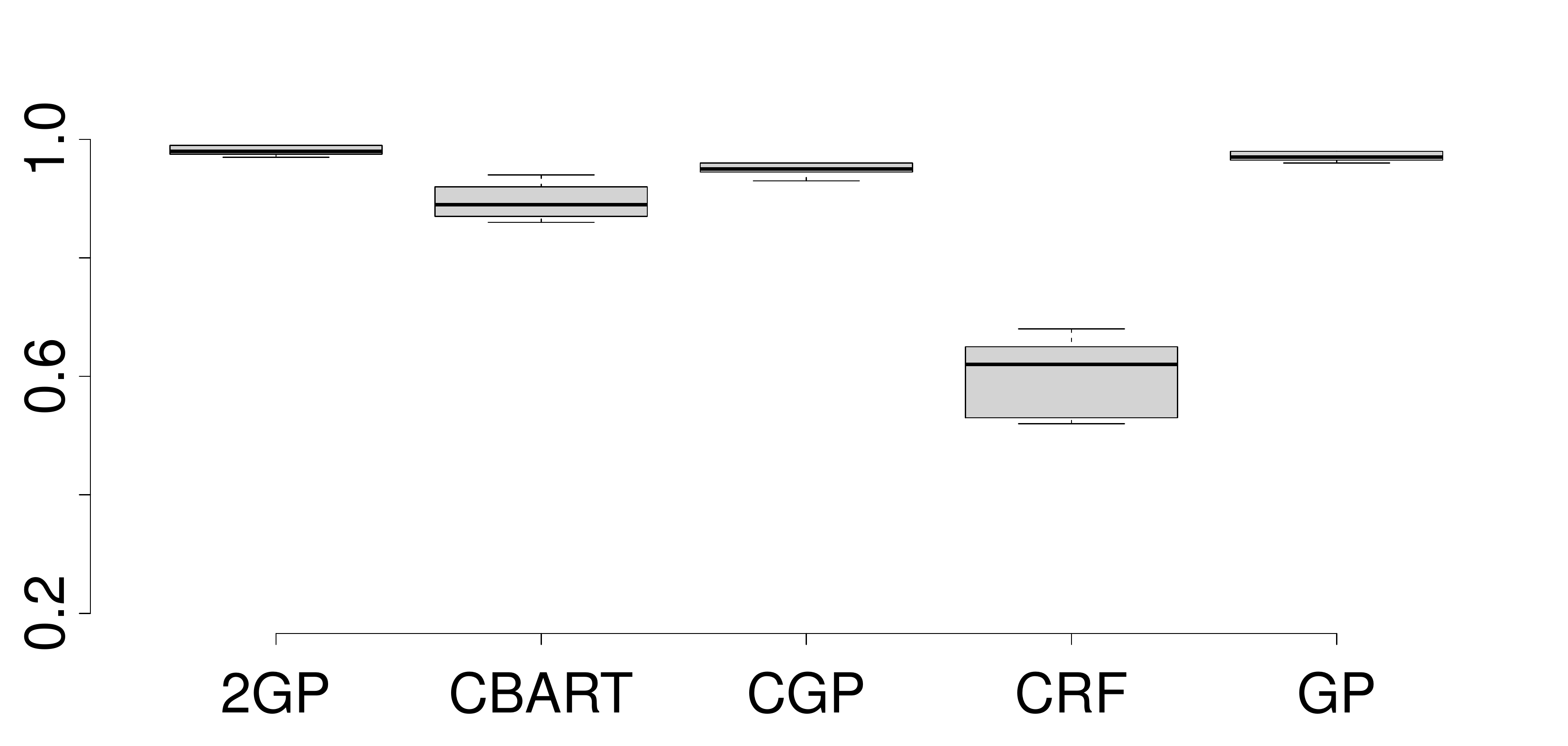}}\\
\caption{coverage of 95\% PI's for CGP, GP, CRF, CBART, 2GP}\label{fig19}
\end{figure}

\begin{figure}[h]
\centering
    \subfigure[$n=5,000$, $\tau=.03$, $p=10,000$]{\label{lengDt}\includegraphics[width=7cm]{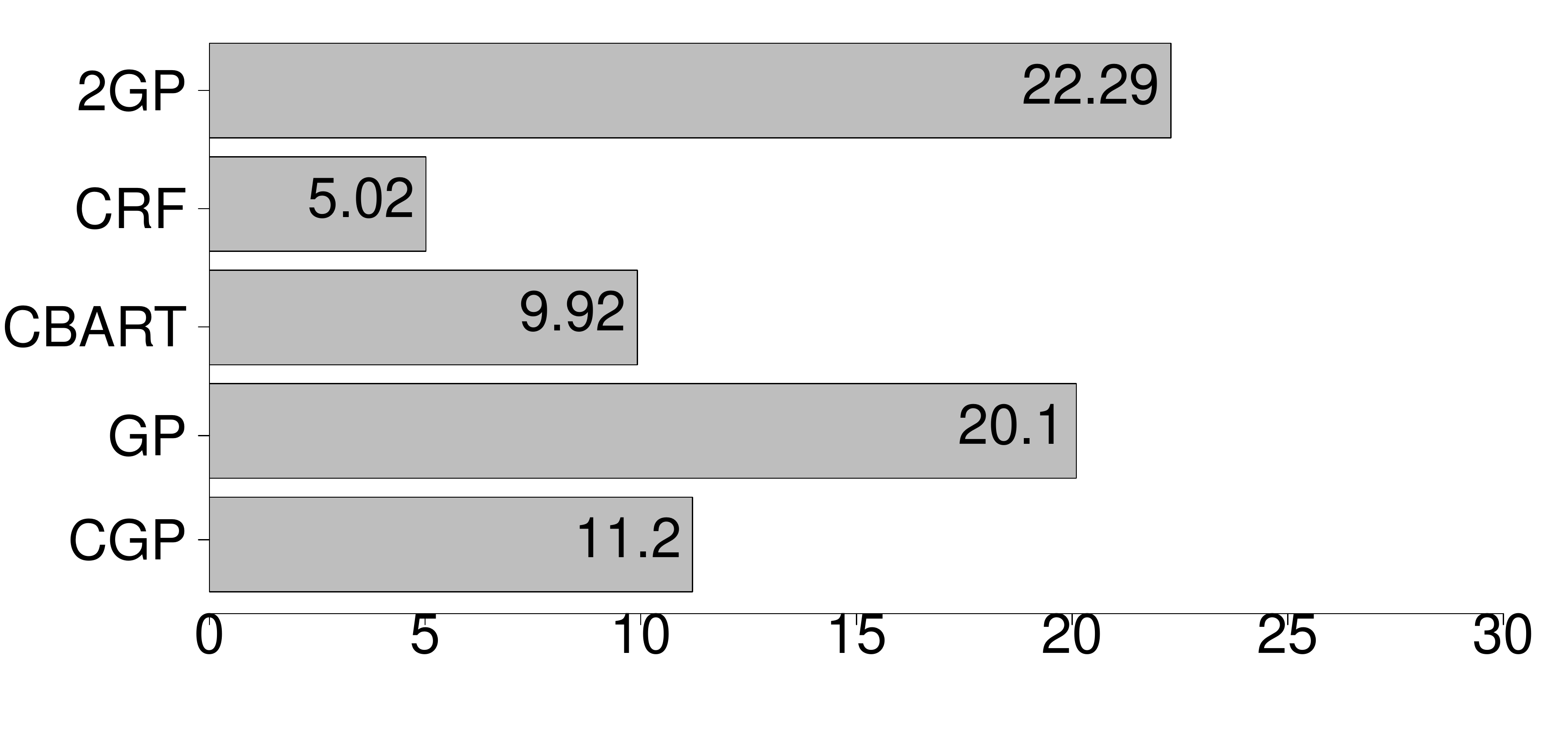}}
    \subfigure[$n=5,000$, $\tau=.03$, $p=20,000$]{\label{lengEt}\includegraphics[width=7cm]{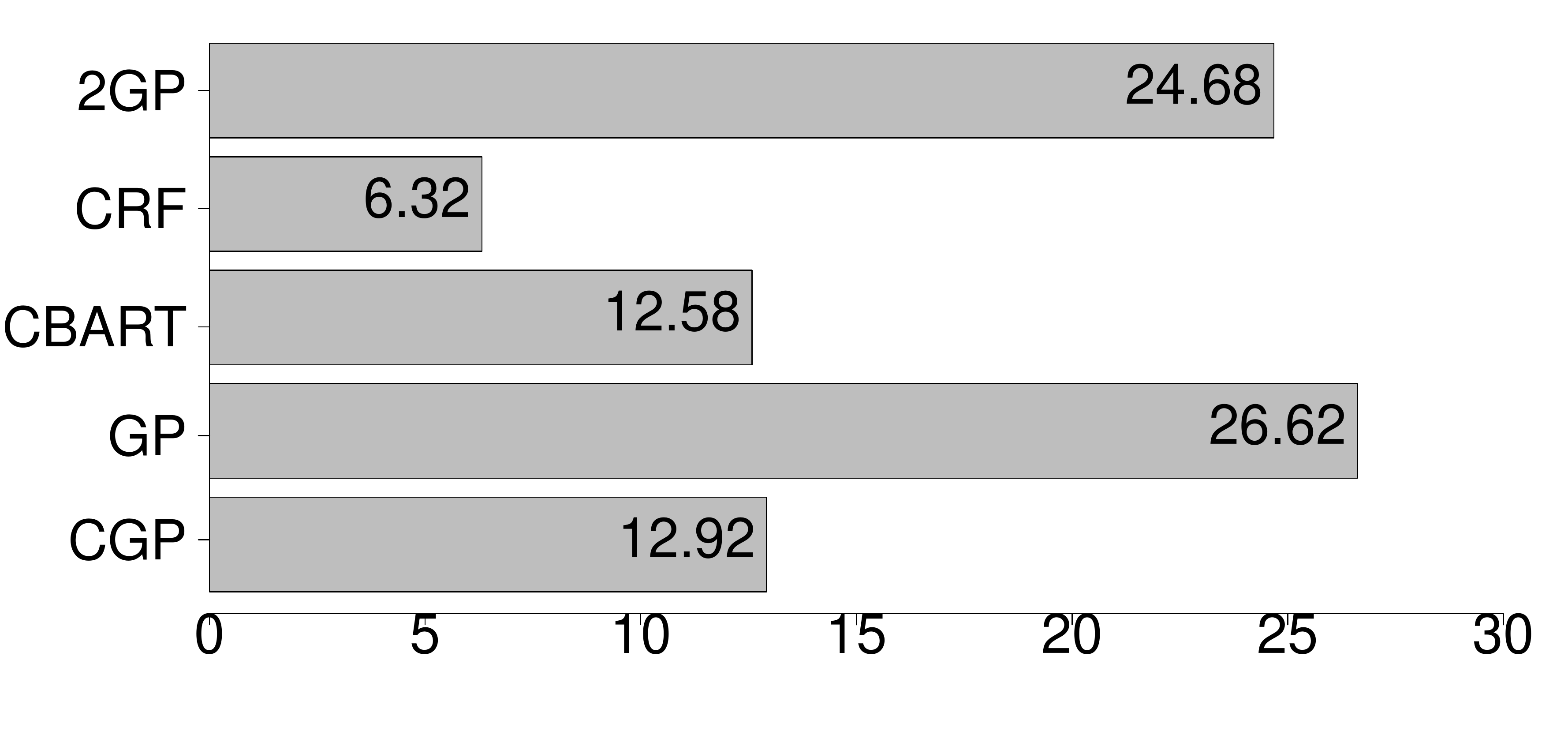}}\\
    \subfigure[$n=5,000$, $\tau=.06$, $p=10,000$]{\label{lengFt}\includegraphics[width=7cm]{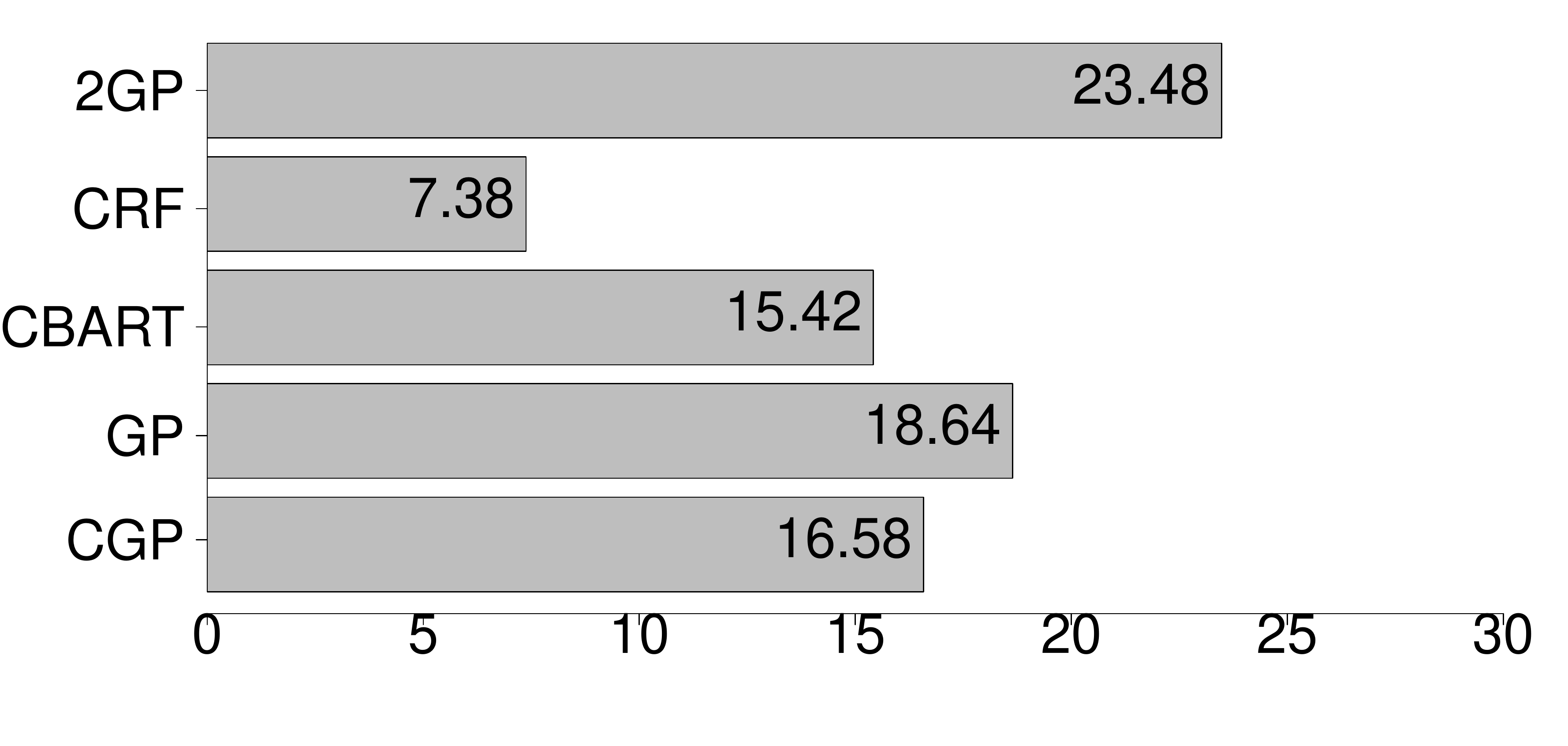}}
    \subfigure[$n=5,000$, $\tau=.06$, $p=20,000$]{\label{lengGt}\includegraphics[width=7cm]{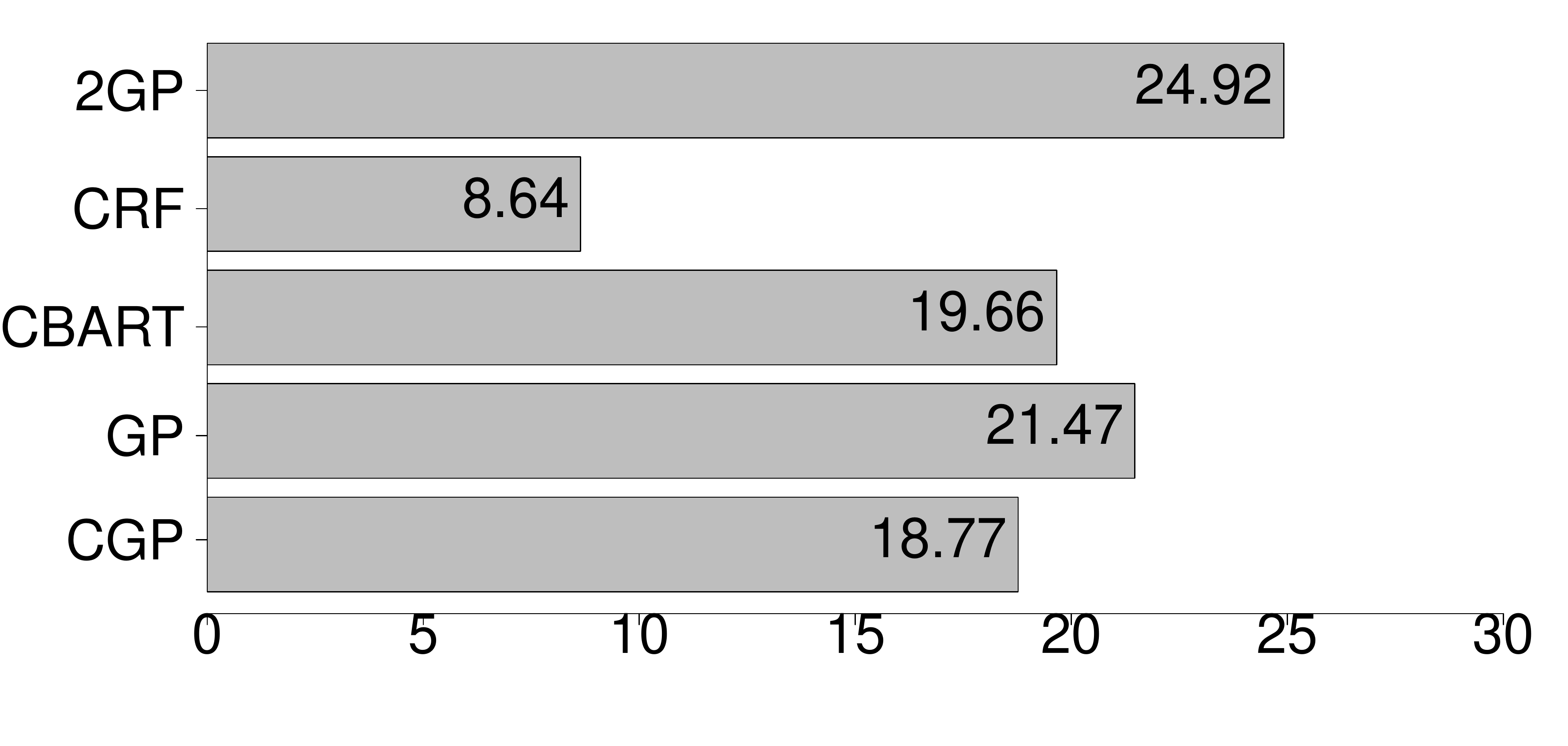}}\\
    \subfigure[$n=5,000$, $\tau=.10$, $p=10,000$]{\label{lengDt}\includegraphics[width=7cm]{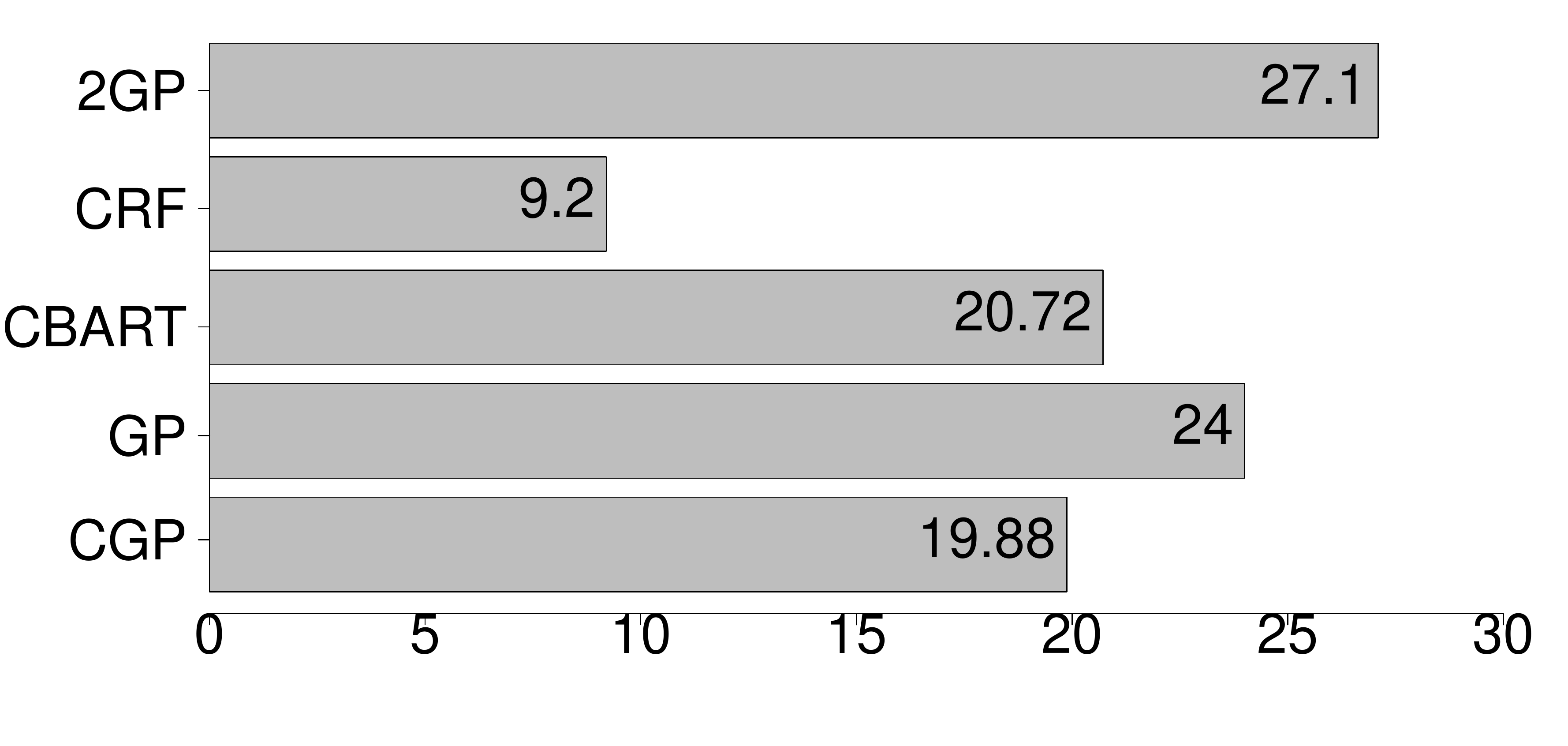}}
    \subfigure[$n=5,000$, $\tau=.10$, $p=20,000$]{\label{lengEt}\includegraphics[width=7cm]{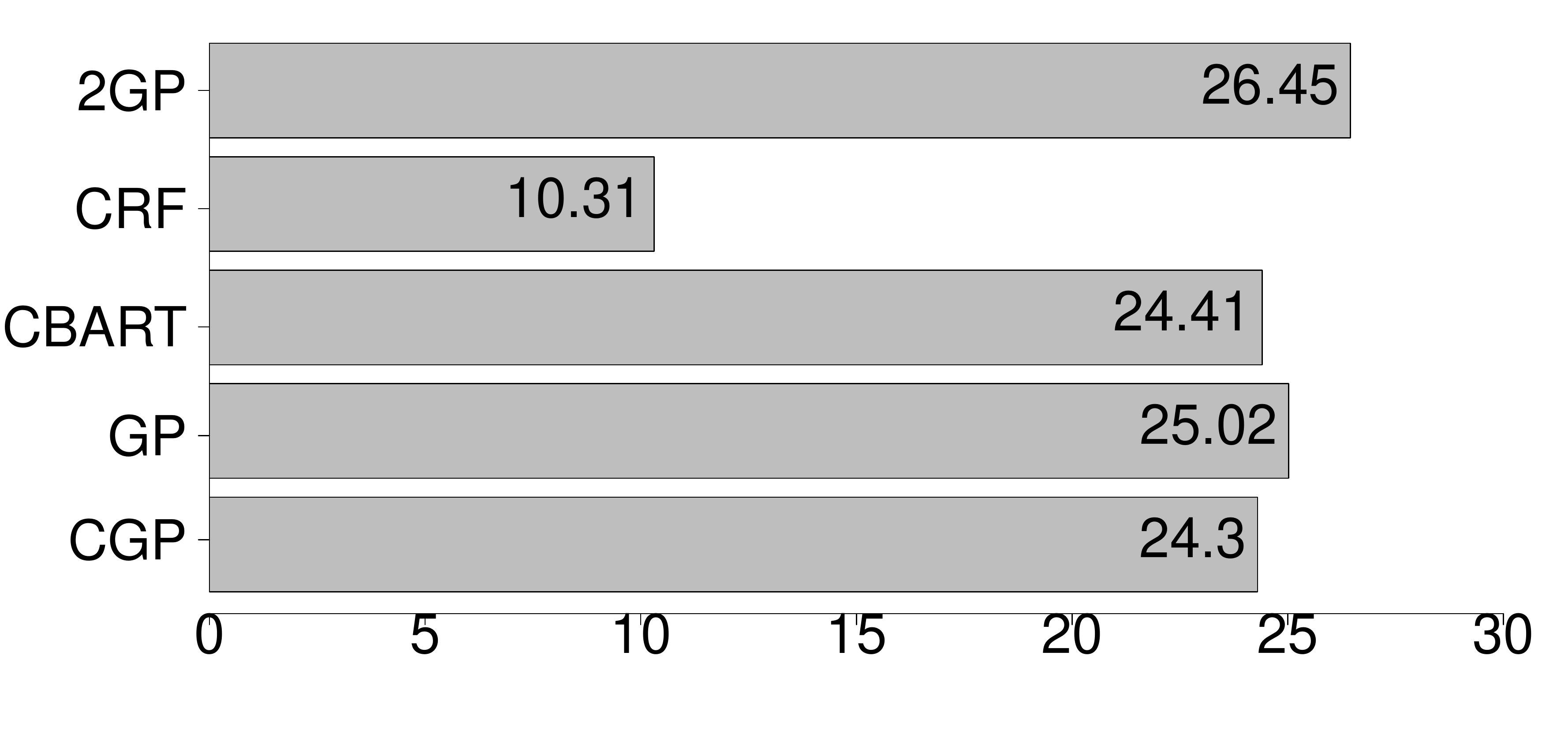}}
\caption{lengths of 95\% PI's for CGP, GP, CBART, CRF, 2GP}\label{fig20}
\end{figure}

In all the simulation scenarios, DSL is the best performer in terms of MSPE, consistent with the routine use of DSL in large scale settings. However, the performance is extremely sensitive to the choice of clusters. In real data applications often inaccurate clustering leads to “suboptimal” performance, as will be seen in the data analysis. Additionally, we are not just interested in obtaining a point prediction approach, but want to obtain methods that provide an accurate characterization of predictive uncertainty.  With this in mind, we additionally examine coverage probabilities and lengths of 95\% predictive intervals (PIs).
\begin{table}[h]
\centering
\begin{tabular}{|  l | l | c | c | c | c| }
\cline{1-5}
 & \multicolumn{4}{ |c| }{Noise in the feature} \\
\cline{2-5}
 &  & .03 & .06 & .10 \\ \hline
\multirow{5}{*}{$p=10,000$} & CGP &  $0.63_{0.038}$ & $1.63_{0.092}$ & $2.31_{0.161}$\\ 
 & GP & $2.02_{0.499}$ & $1.70_{0.287}$ & $3.07_{0.360}$ \\
 & CRF & $0.92_{0.070}$ & $2.20_{0.176}$ & $3.44_{0.136}$ \\
 & CBART & $0.79_{0.072}$ & $1.58_{0.111}$ & $2.74_{0.068}$ \\
 & DSL   & $0.44_{0.023}$ & $0.45_{0.015}$ & $0.50_{0.035}$ \\
 & 2GP   & $3.80_{0.481}$ & $4.05_{0.434}$ & $4.10_{0.350}$ \\
 \hline
\multirow{5}{*}{$p=20,000$} & CGP &  $1.24_{0.042}$ & $2.01_{0.104}$ & $3.48_{0.217}$ \\ 
 & GP & $2.31_{0.418}$ & $2.23_{0.323}$ & $3.29_{0.330}$ \\
 & CRF & $1.62_{0.070}$ & $2.99_{0.224}$ & $4.21_{0.224}$ \\
 & CBART & $1.22_{0.082}$ & $2.59_{0.146}$ & $3.84_{0.191}$ \\
 & DSL & $0.48_{0.024}$ & $0.47_{0.016}$ & $0.53_{0.080}$ \\
 & 2GP & $3.93_{0.592}$ & $4.10_{0.372}$ & $4.55_{0.481}$ \\
\hline
\end{tabular}
\caption{$MSPE\times 0.1$ along with the bootstrap $sd\times 0.1$ for all the competitors}\label{tab12}
\end{table}
Boxplots for coverage probabilities of 95\% PI's  are presented in figure~\ref{fig19}. Figure~\ref{fig20} presents lengths of 95\% prediction intervals for all the competitors.
As expected, CGP, GP, 2GP and CBART demonstrate better performance in terms of coverage. However, in low noise cases CGP and CBART achieve similar coverage with a two fold reduction in the length of PIs compared to GP or 2GP. CRF, like in the previous section, shows under-coverage with narrow predictive intervals. The predictive interval for CGP is found to be a marginally wider than CBART with comparable coverage. With high noise features tend to lose their manifold structure more and hence performance is affected for all the competitors. It is observed that with high noise all of them tend to have wider predictive intervals. DSL presents overly narrow predictive intervals (not shown here) yielding severe under-coverage.


\subsection{Computation Time}

 One of the major motivations in developing CGP was to massively improve computational scalability to huge $p$, $n$ settings. Clearly, the computational time for nonparametric estimation methods such as BART, TGP or RF applied to the original data will become notoriously prohibitive for huge $p$, and hence we focus on comparisons with more scalable methods. The approach of applying BART, RF and TGP to the compressed features, which is employed in CBART, CRF and CTGP respectively, is faster to implement. Using non-optimized R code implemented on a single 3.06-GHz
Intel Xeon processor with 4.0 Gbytes of random-access memory running Debian LINUX, the computing time for 2,000 iterations of CBART in the $n=100$ and $p=10,000, 20,000$ are only $7.21, 8.36$ seconds, while CGP has run time of $7.48, 8.05$ seconds, respectively. Increasing $n$ moderately we find CBART and CGP have similar run time. CRF is a bit faster than both of them, while CTGP has run time $37.64, 38.33$ seconds for $p=10,000, 20,000$ respectively. For moderate n, 2GP is found to have similar run time as CBART.

With huge $n$, CTGP is impractically slow to run and hence it is omitted in the comparison. GP needs to calculate and store distance matrix of $p$ features. Apart from the storage bottleneck, such a procedure incurs a computational complexity of $O(n^2p)$ which is massive for large $n,p$.
 CGP instead proposes calculating and storing a distance matrix of $m$ compressed features, with a computational complexity of $O(n^2m)$. computation time for CGP additionally depends on a number of factors, (i) Gram Schmidt orthogonalization of $m$ rows of $m\times p$ matrices, (ii) inverting an $m_{\bPhi}\times m_{\bPhi}$ matrix for large n, (iii) multiplying $n\times p$ and $p\times m$ matrices. For large $n$ along with these three steps, one requires multiplying $n\times m_{\bPhi}$ with
$m_{\bPhi}\times m_{\bPhi}$. All of these steps are computationally less demanding even with large $n$ and $p$. The computation is further facilitated by the easy parallelizing over different choices of $m$ in the model averaging step. Even not exploiting any parallelization, one obtains results quickly using non-optimized R code on a single 3.06-GHz Intel Xeon processor with 4.0 Gbytes of random-access memory running Debian LINUX.

\begin{figure}[h!]
\centering
    \subfigure[$n=1,000$]{\label{comp1}\includegraphics[width=10cm]{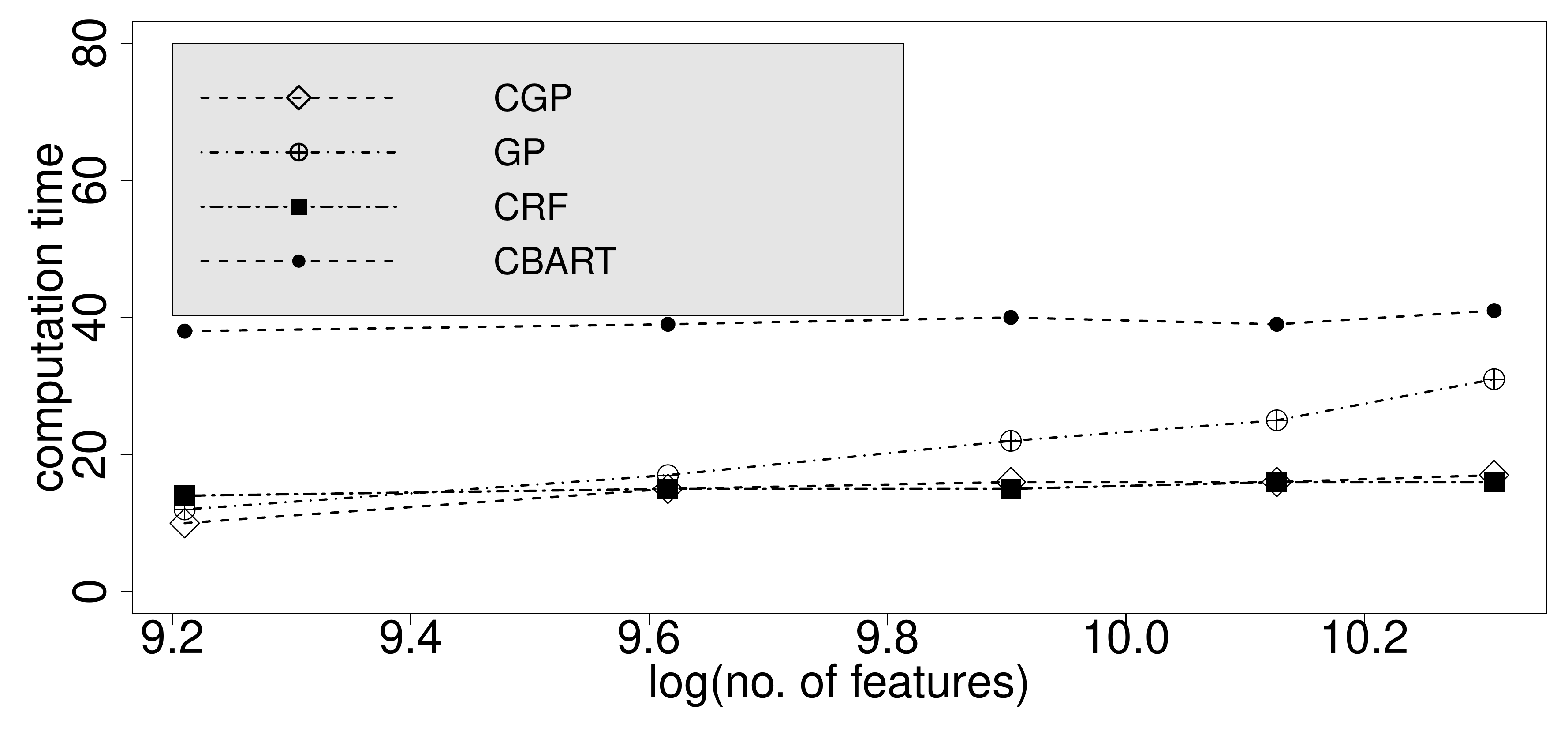}}\\
    \subfigure[$n=3,000$]{\label{comp2}\includegraphics[width=10cm]{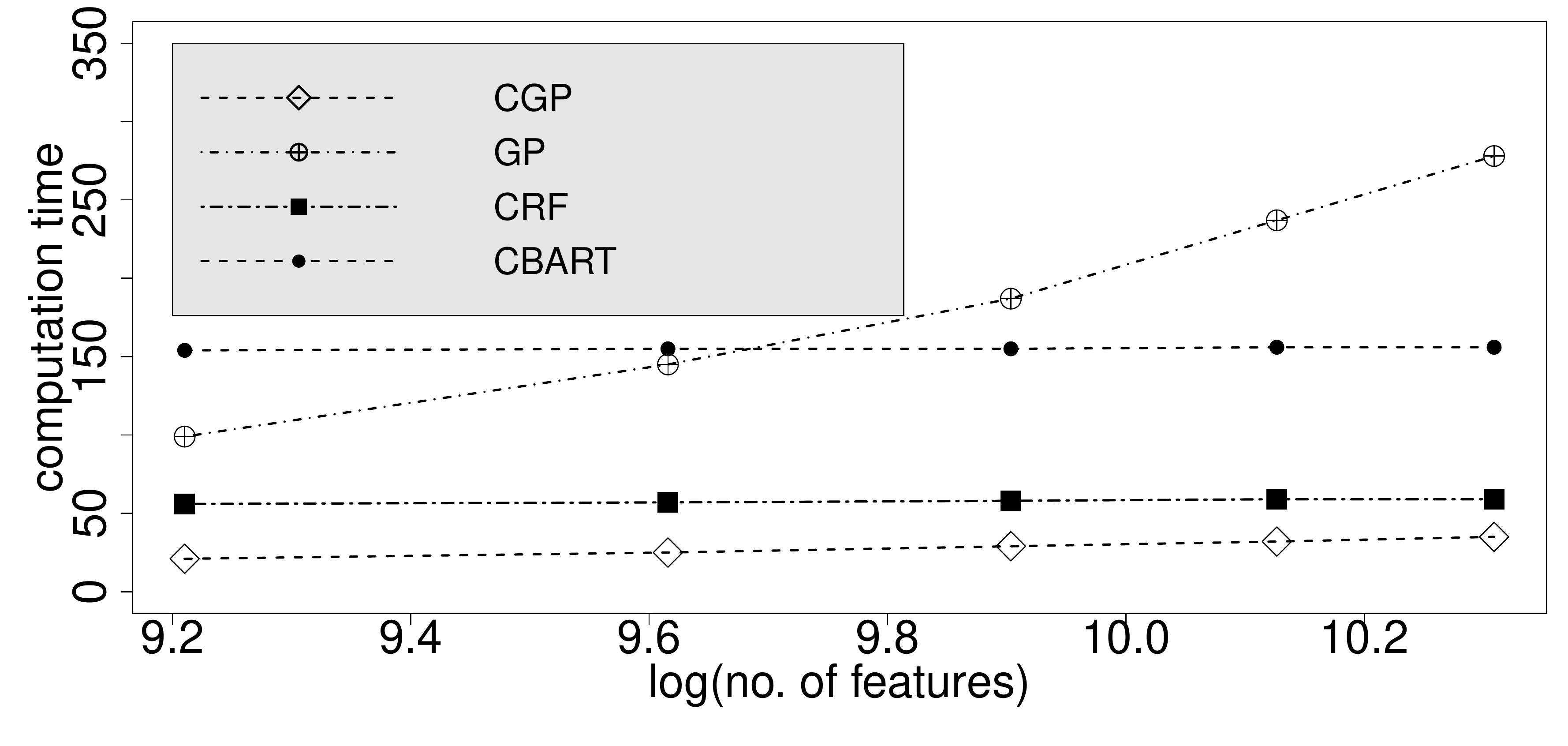}}\\
    \subfigure[$n=5,000$]{\label{comp2}\includegraphics[width=10cm]{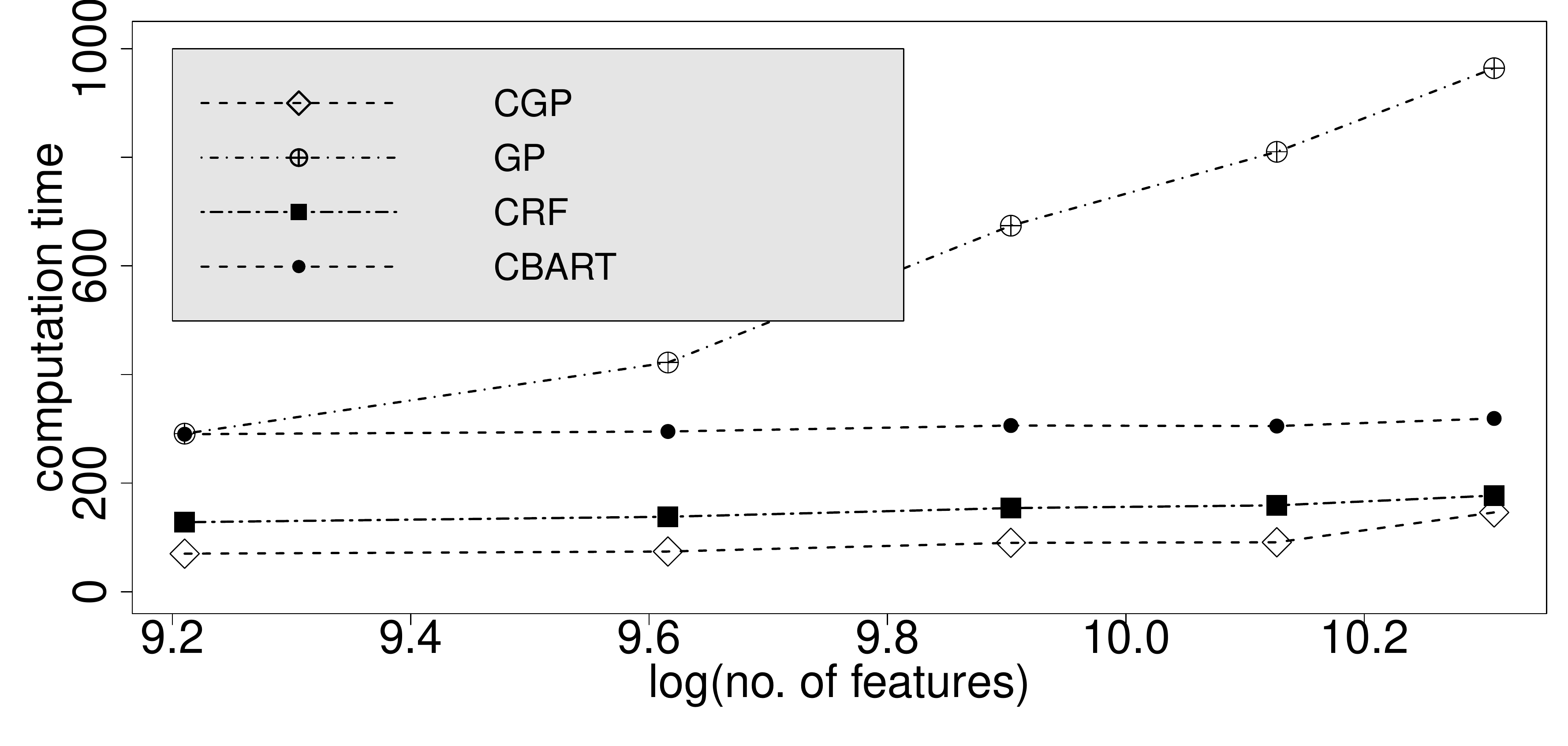}}
\caption{log of the Computational time in seconds for CGP, GP, CBART, CRF against log of the number of features.}\label{timecomp}
\end{figure}

Figure~\ref{timecomp} shows the computational speed comparison between CGP, GP, CBART and CRF for various $n$ and $p$. Computational speed is recorded assuming existence of a number of processors on which parallelization can be executed, whenever needed. Clearly, as $n$ increases, CGP enjoys substantial computational advantage over all its competitors. The computational advantage is especially notable over CBART and GP. Run times of DSL are also recorded for $n=5,000$ and $p=10,000,15,000, 20,000, 25,000, 30,000$ and they turn out to be $449, 599, 737, 945, 1158$ seconds respectively. On the other hand, 2GP involves creating adjacency matrices followed by eigen-decomposition of $n\times n$ matrix. Both these steps are computationally demanding. It has been observed that 2GP takes $602, 723, 856, 983, 1108$ seconds to run for $n=5000$ and $p=10000, 15000, 20000, 25000, 30000$ respectively. Therefore, CGP can outperform even a simple two stage estimation procedure such as DSL in terms of computational speed.
Such rapid computation offered by CGP becomes particularly notable as we scale from tens of thousands of samples and features to millions or more, which is becoming increasingly common.

\section{Application to Face Images}

\begin{figure}[h!]
\centering
    \subfigure{\label{f1}\includegraphics[width=4cm]{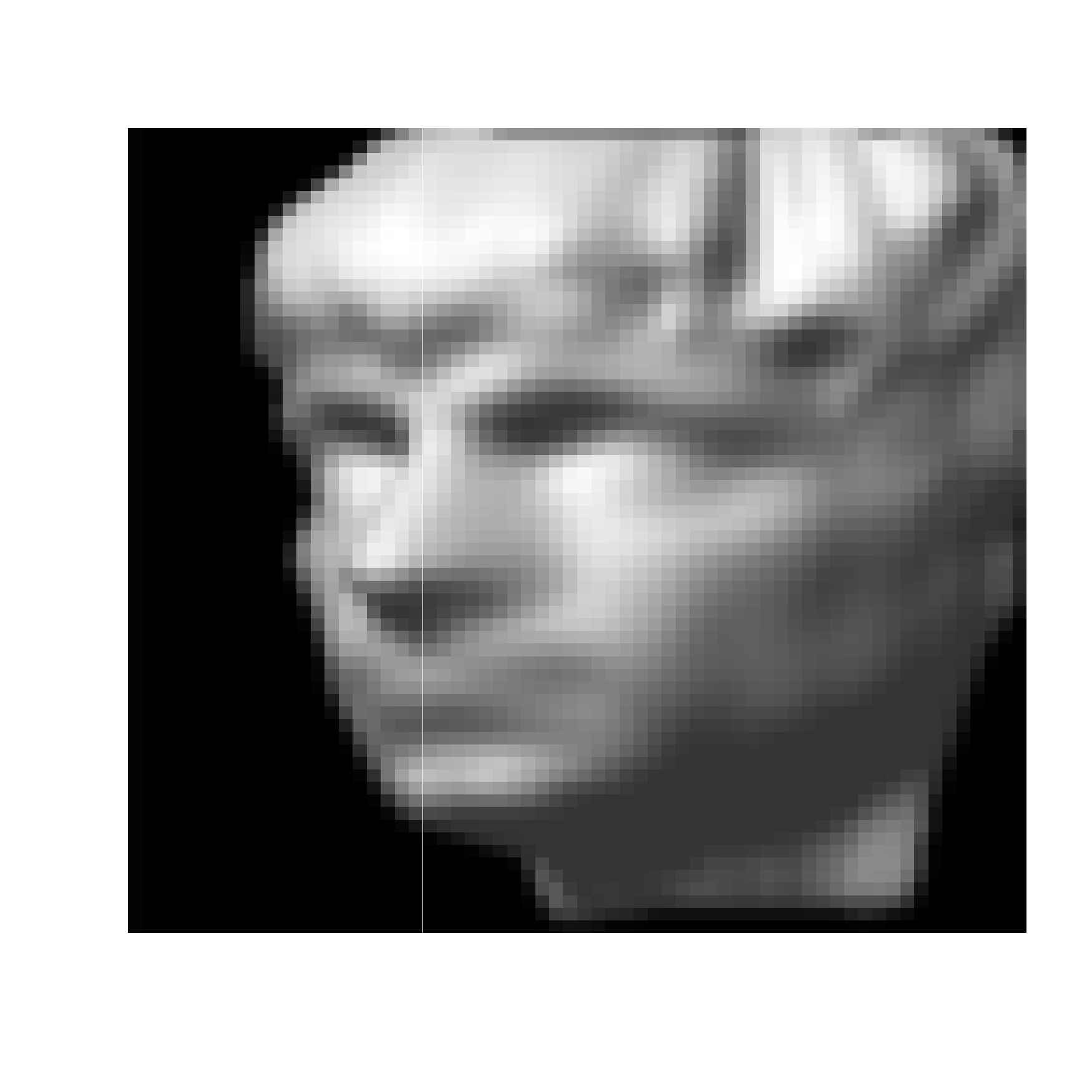}}
    \subfigure{\label{f2}\includegraphics[width=4cm]{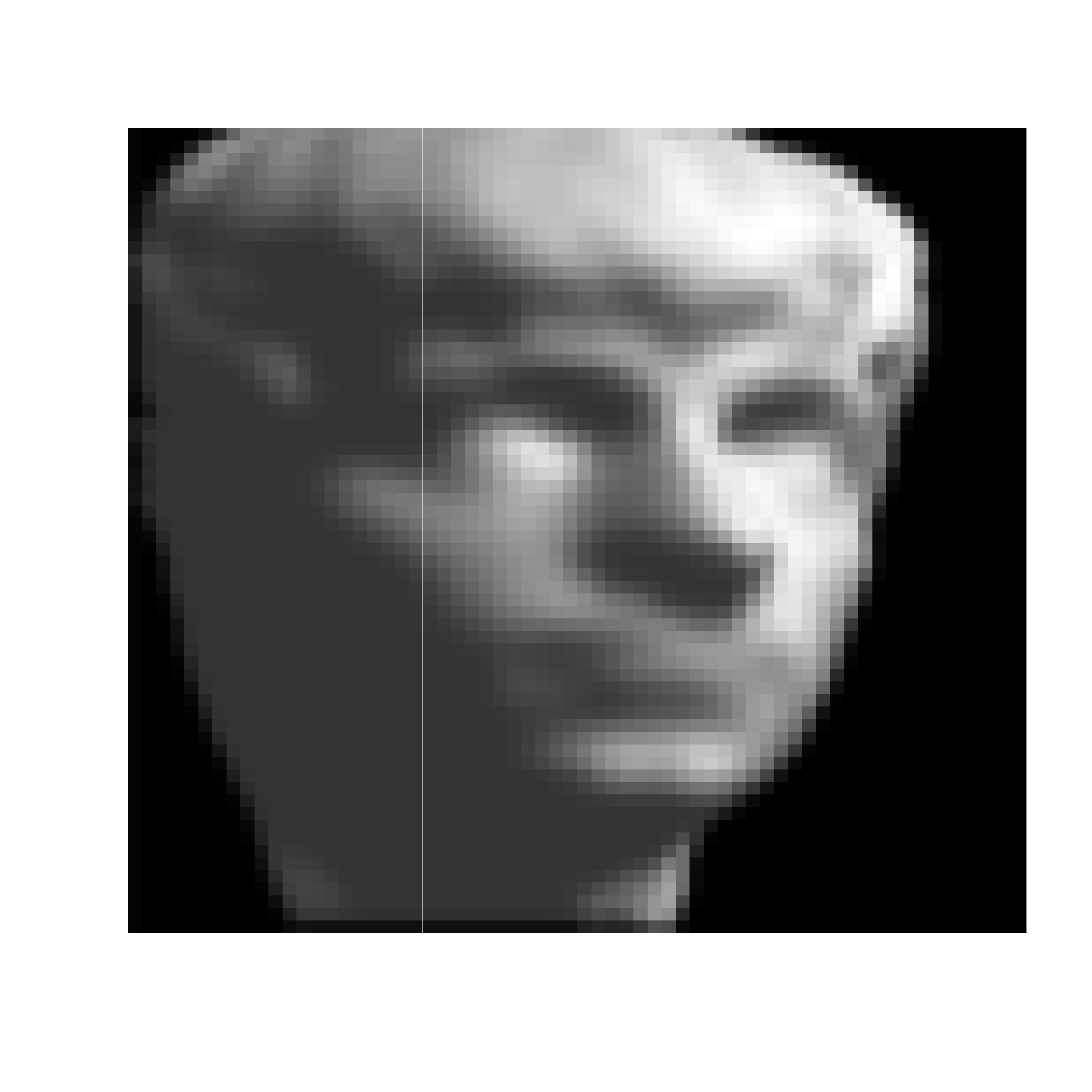}}
    \subfigure{\label{f3}\includegraphics[width=4cm]{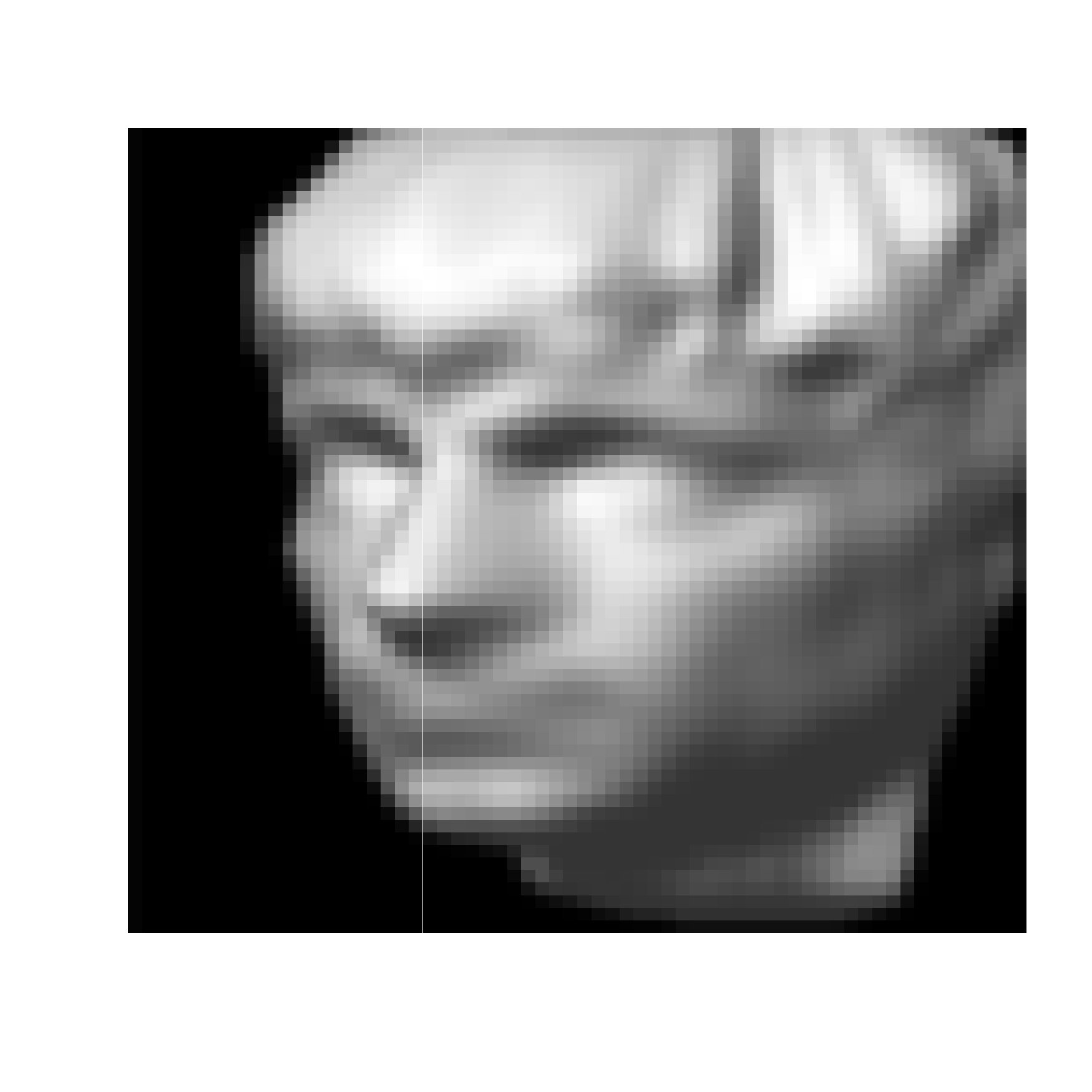}}\\
    \subfigure{\label{f4}\includegraphics[width=4cm]{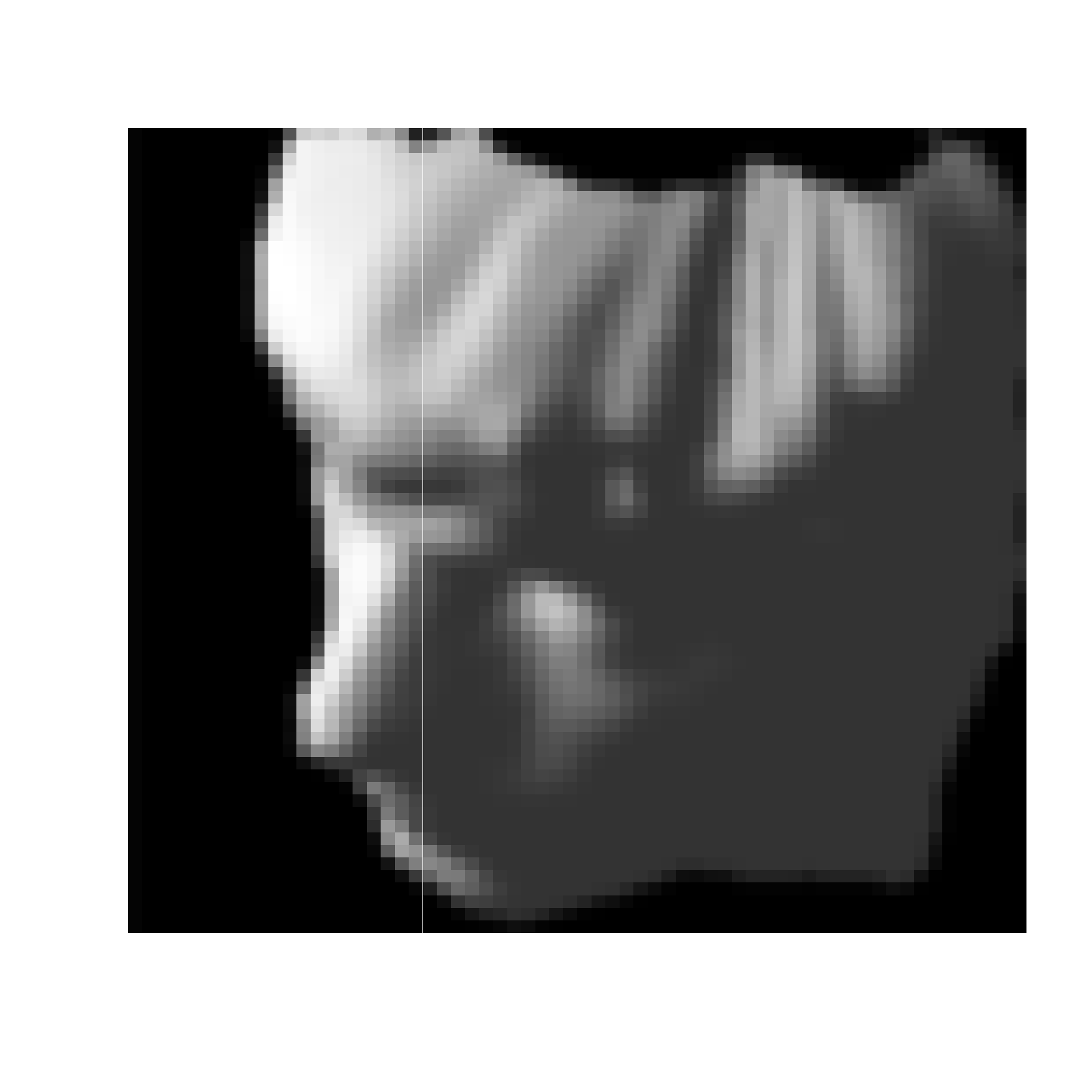}}
    \subfigure{\label{f6}\includegraphics[width=4cm]{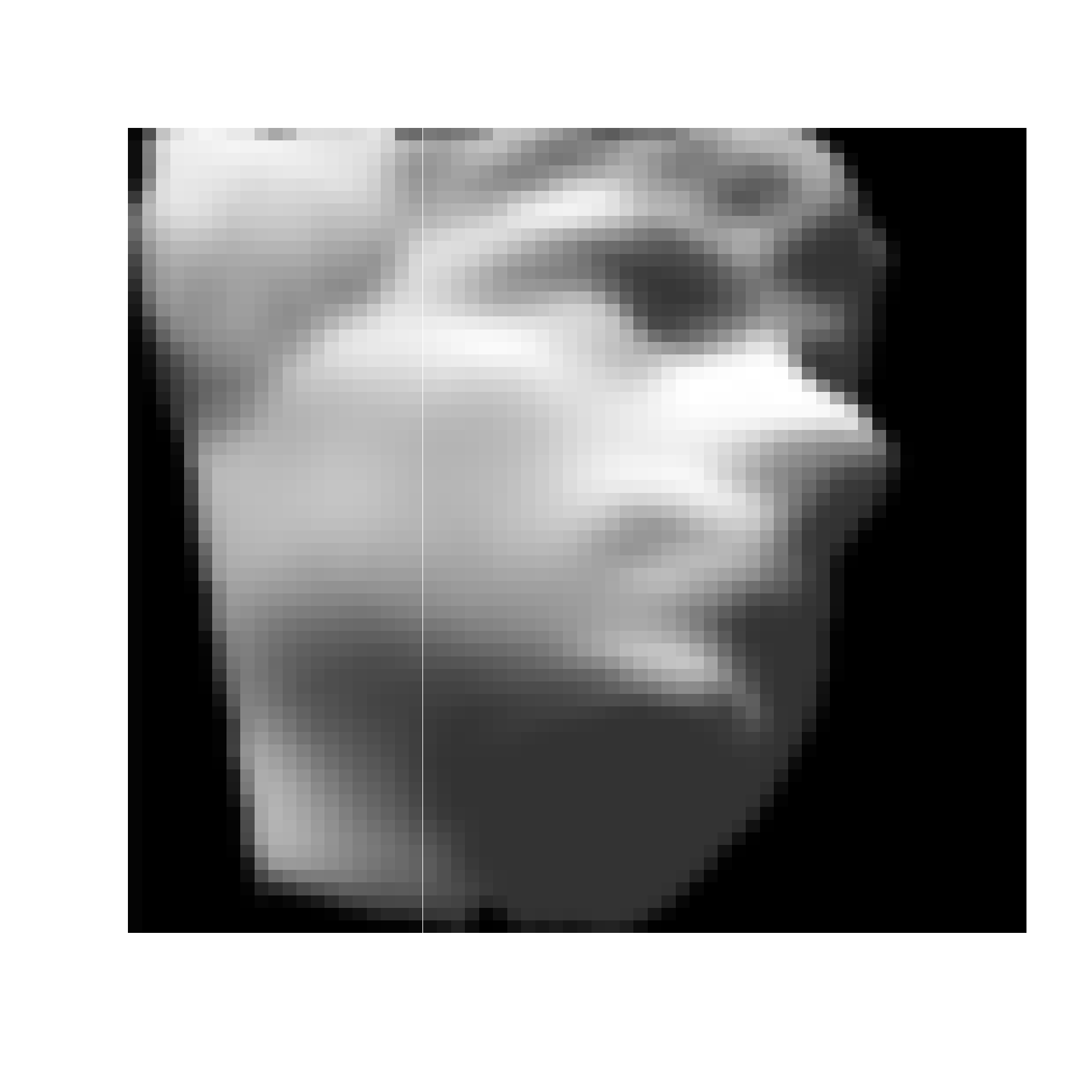}}
    \subfigure{\label{f8}\includegraphics[width=4cm]{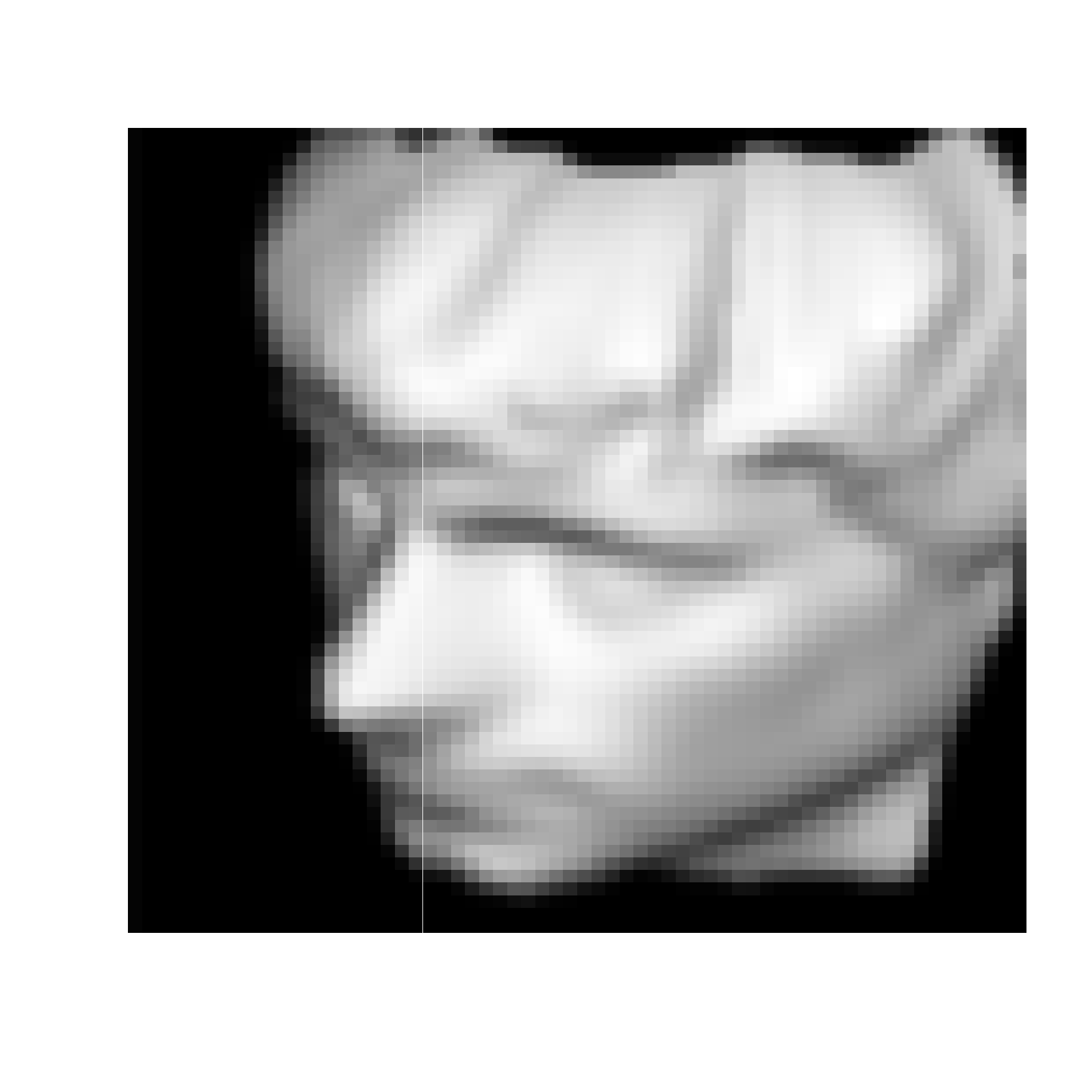}}
\caption{Representative images from the Isomap face data.}\label{faces}
\end{figure}
In our simulation examples, the underlying manifold is three dimensional and can be directly visualized. In this section we present an application in which both the dimension and the structure of the underlying manifold is unknown. The dataset consists of 698 images of an artificial face and is referred to as the \emph{Isomap face data} (\cite{tenenbaum2000global}). A few such representative images are presented in Figure~\ref{faces}. Each image is labeled with three different variables: illumination-, horizontal- and vertical-orientation. Three dimensional images observed in the data should ideally form three dimensional manifold. However, only a two dimensional projection of the images are presented in the data in the form of matrices of the order $64\times 64$ pixels in size. intuitively a limited number of additional features are needed for different views of the face. This is confirmed by the recent work of \cite{levina2004maximum,aswani2011regression} where the intrinsic dimensionality is estimated to be small from these images. More details about the dataset can be found in \url{http://isomap.stanford.edu/datasets.html}.

We apply CGP and all the competitors to the dataset to assess relative performances. To set up the regression problem, we consider horizontal pose angles (vary in $[-75^0,75^0]$) of the images, after standardization, as the responses. The features are taken $64\times 64=4096$ dimensional vectorized images for each sample. To deal with more realistic situations, $N(0,\tau^2)$ noise is added to each pixel of the images, with varying $\tau$, to make predictive inference more challenging from the noisy images. We carry out random splitting of the data into $n=648$ training cases and $n_{pred}=50$ test cases and run all the competitors to obtain predictive inference in terms of MSPE, length and coverage of 95\% predictive intervals. To avoid spurious inference due to small validation set, this experiment is repeated 20 times.

Table~\ref{tab3:datamspe} presents MSPE for all the competing methods averaged over 20 experiments along with their standard errors computed using 100 bootstrap samples.
{\footnotesize
\begin{table}[h]
\begin{center}
\begin{tabular}{|c|c|c|c|c|c|c|}
	\hline
$\tau$ & CGP & GP & CBART & CRF & DSL & 2GP\\
	\hline
$0.03$ & $0.14_{0.059}$  & $0.92_{0.074}$ & $0.06_{0.005}$ & $0.05_{0.007}$ & $0.68_{0.023}$ & $0.95_{0.062}$ \\
$0.06$ & $0.09_{0.006}$  & $0.79_{0.056}$ & $0.09_{0.007}$ & $0.09_{0.008}$ & $0.75_{0.015}$ & $0.94_{0.041}$\\
$0.10$ & $0.12_{0.008}$ & $0.83_{0.077}$ & $0.12_{0.005}$ & $0.13_{0.011}$ & $0.54_{0.014}$ & $0.92_{0.013}$\\
	\hline
\end{tabular}
\end{center}
\caption{MSPE and standard error (computed using 100 bootstrap samples) for all the competitors over 20 replications}
\label{tab3:datamspe}
\end{table}}
It is clear from Table~\ref{tab3:datamspe} that CGP along with its compressed competitors explain a lot of variation in the response. GP and 2GP are the worst performers in terms of MSPE. DSL also performs much worse than the compressed competitors.  This is consistent with our experience that, in the presence of a complex and unknown manifold structure along with noise, DSL can be unreliable relative to CGP which tends to be more robust to the type of manifold and noise level.

Note that, because of the standardization, the null model yields MSPE 1. Therefore, it is clear from Table~\ref{tab3:datamspe} that CGP along with its compressed competitors explain a lot of variation in the response. GP and 2GP are the worst performers in terms of MSPE. DSL also performs much worse than the compressed competitors.  This is consistent with our experience that, in the presence of a complex and unknown manifold structure along with noise, DSL can be unreliable relative to CGP which tends to be more robust to the type of manifold and noise level.

\begin{figure}[h!]
\centering
    \subfigure[$\tau=0.03$]{\label{comp1}\includegraphics[width=7cm]{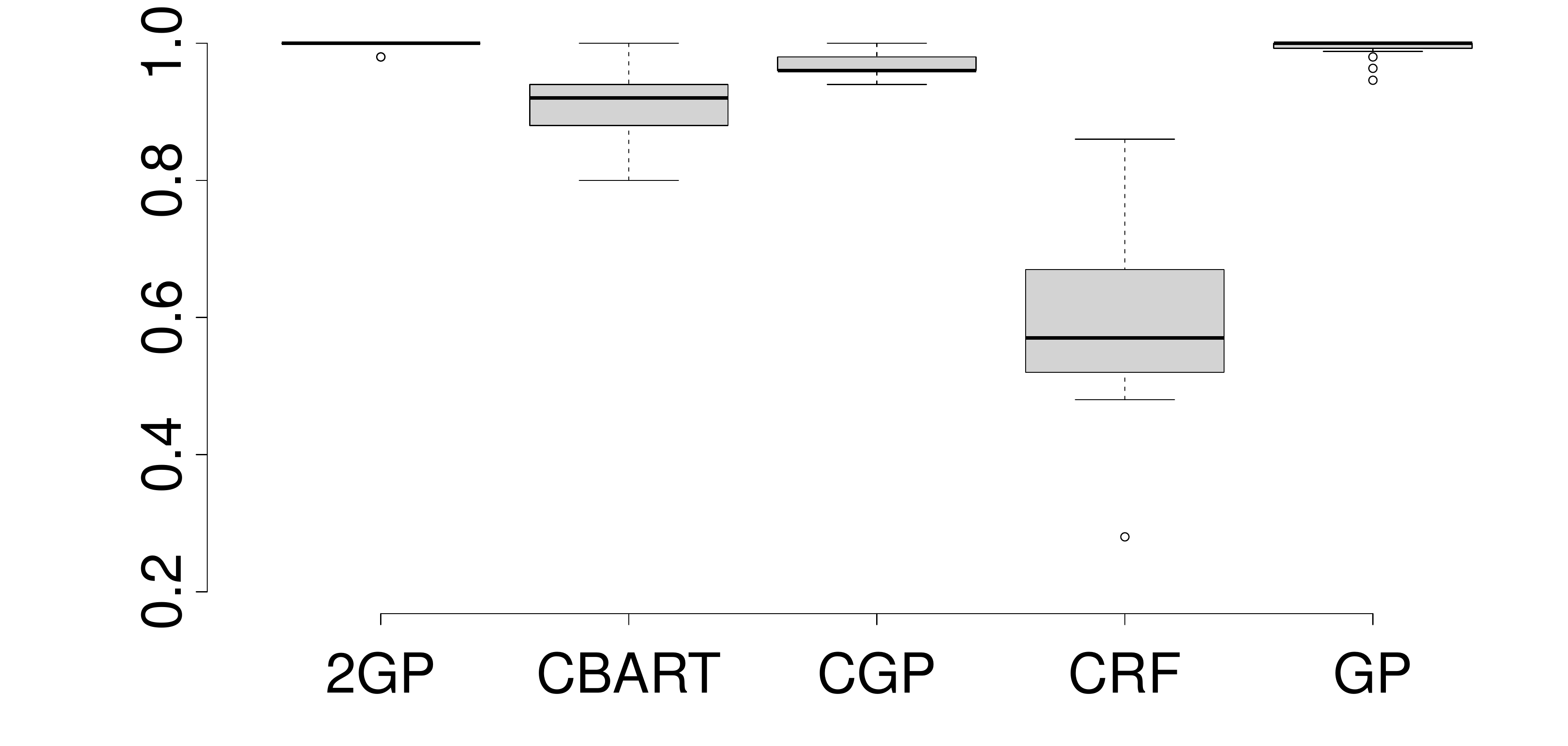}}
    \subfigure[$\tau=0.03$]{\label{comp2}\includegraphics[width=7cm]{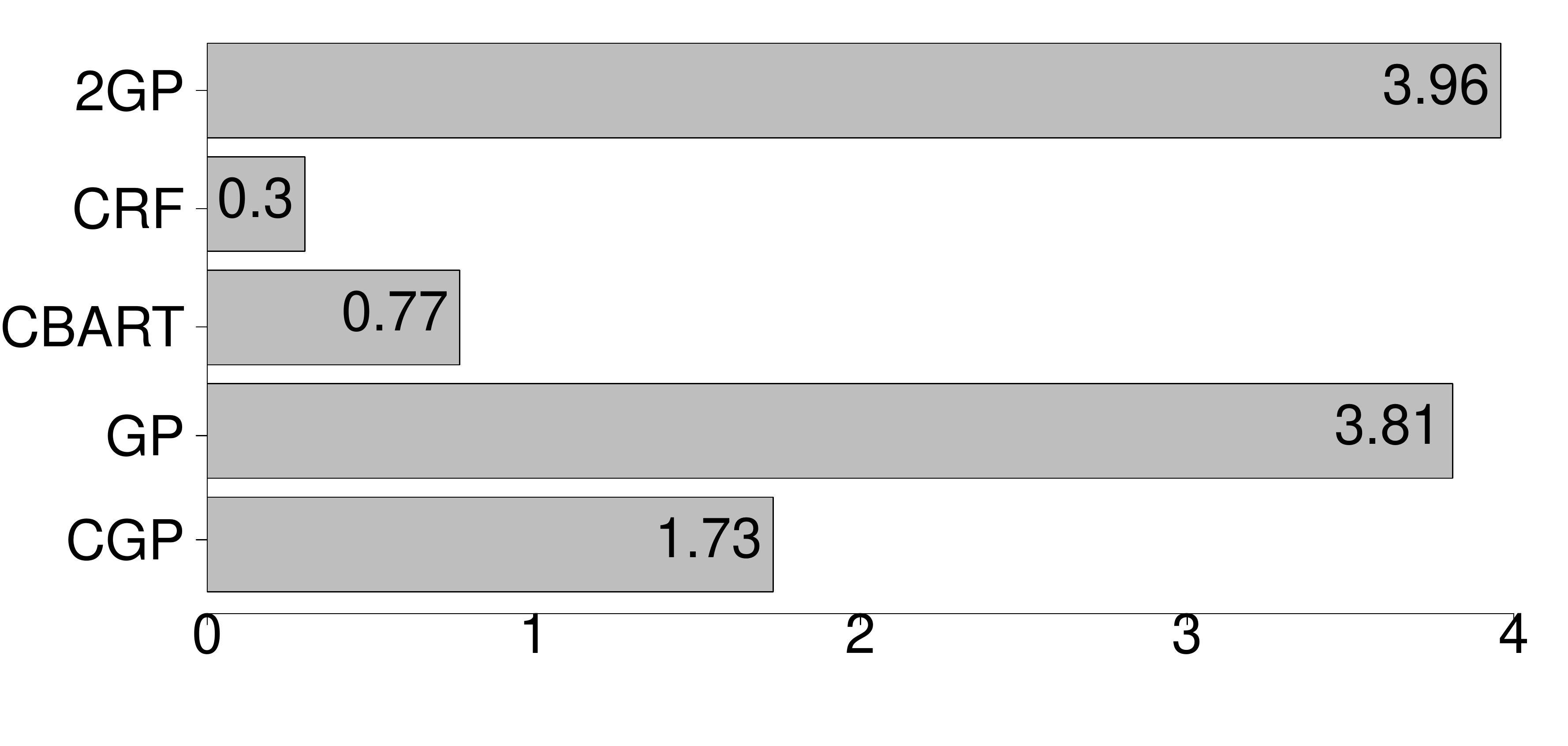}}\\
    \subfigure[$\tau=0.06$]{\label{comp3}\includegraphics[width=7cm]{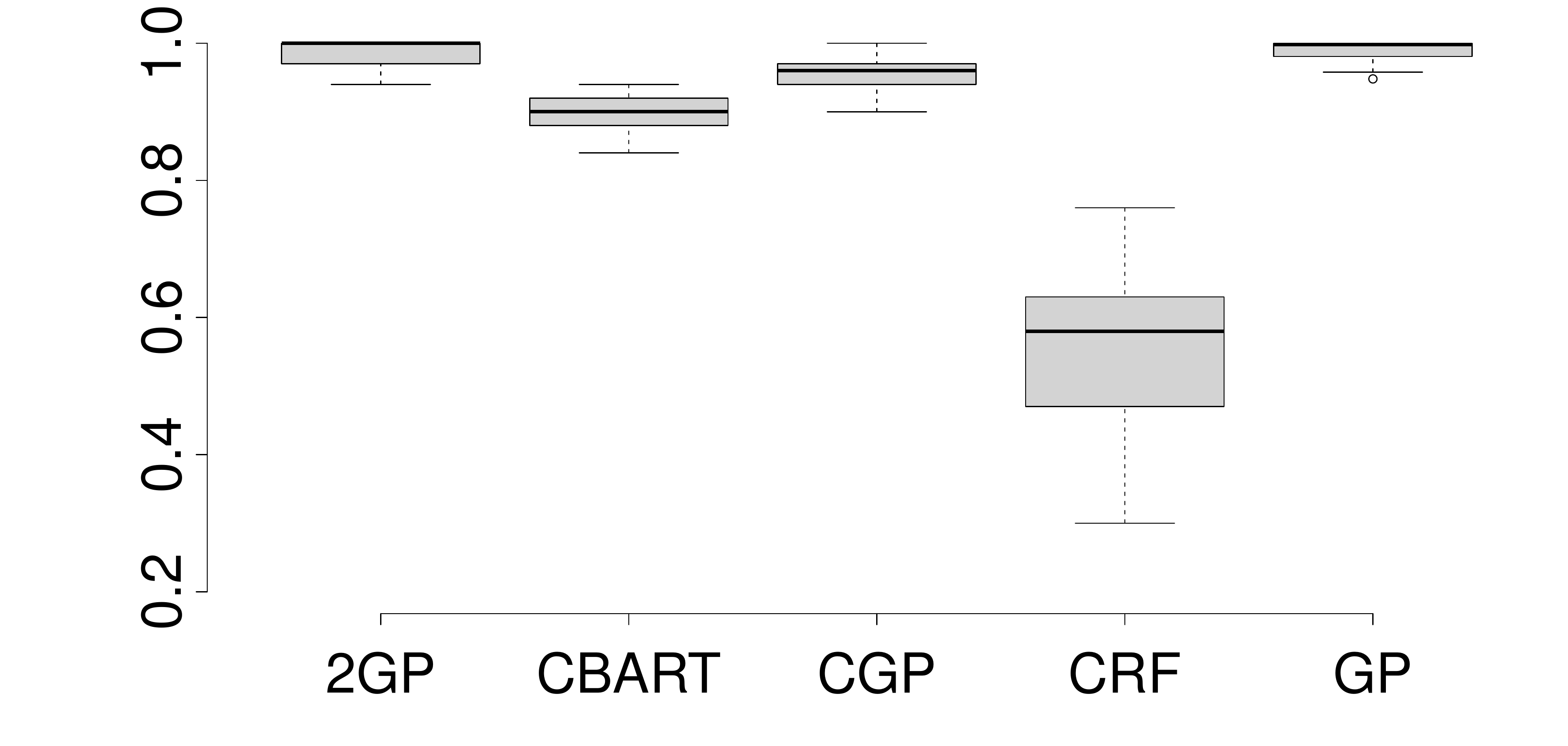}}
    \subfigure[$\tau=0.06$]{\label{comp4}\includegraphics[width=7cm]{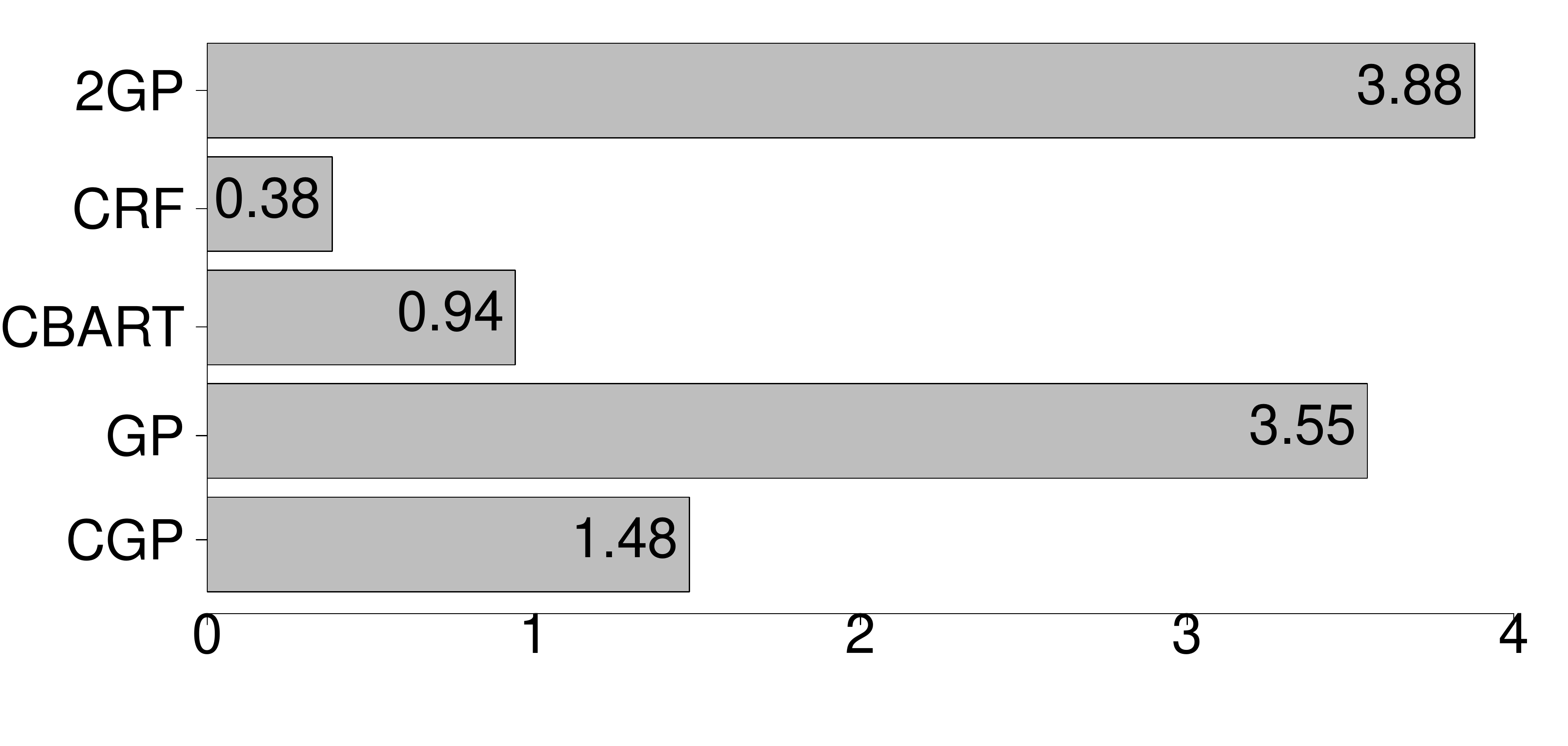}}\\
    \subfigure[$\tau=0.10$]{\label{comp5}\includegraphics[width=7cm]{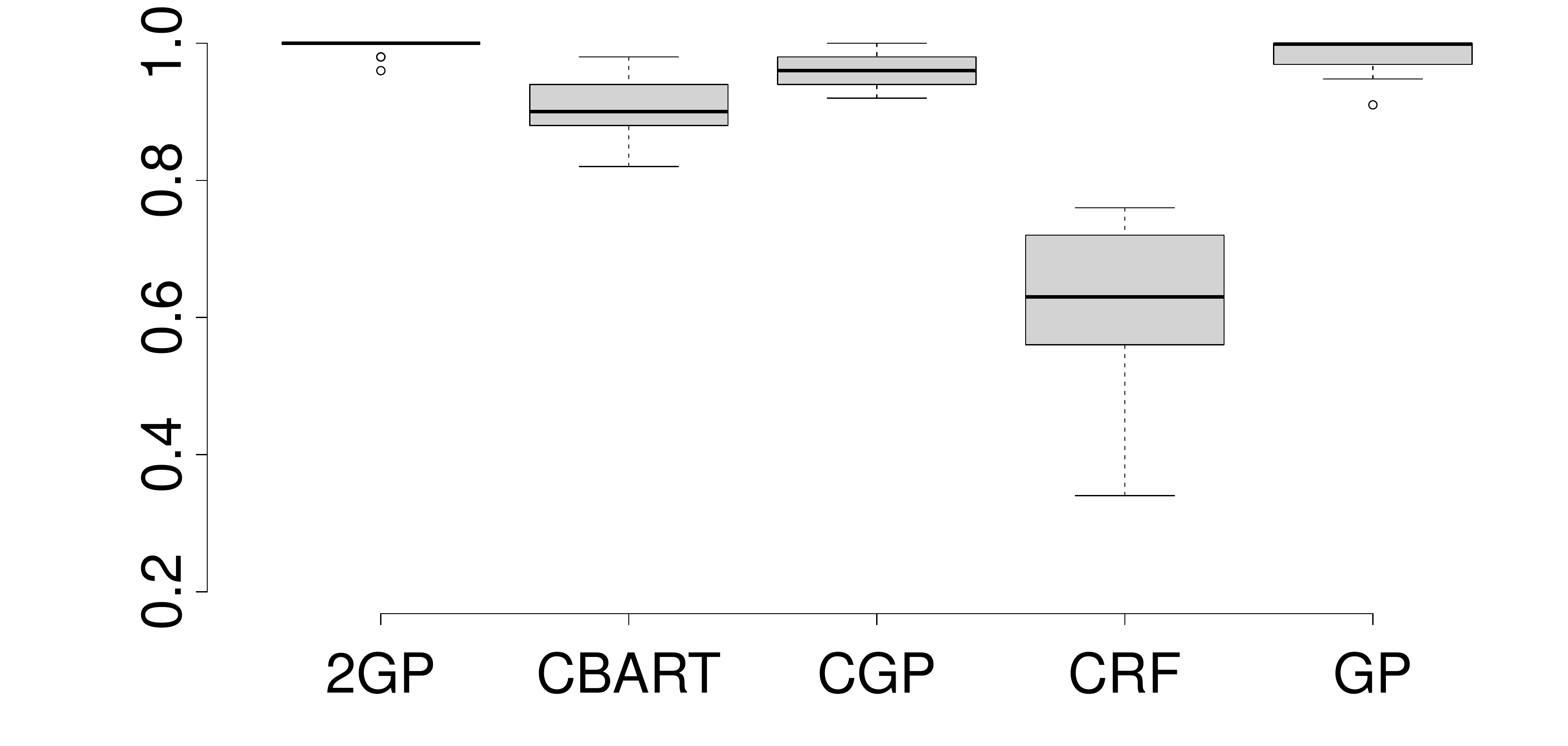}}
    \subfigure[$\tau=0.10$]{\label{comp6}\includegraphics[width=7cm]{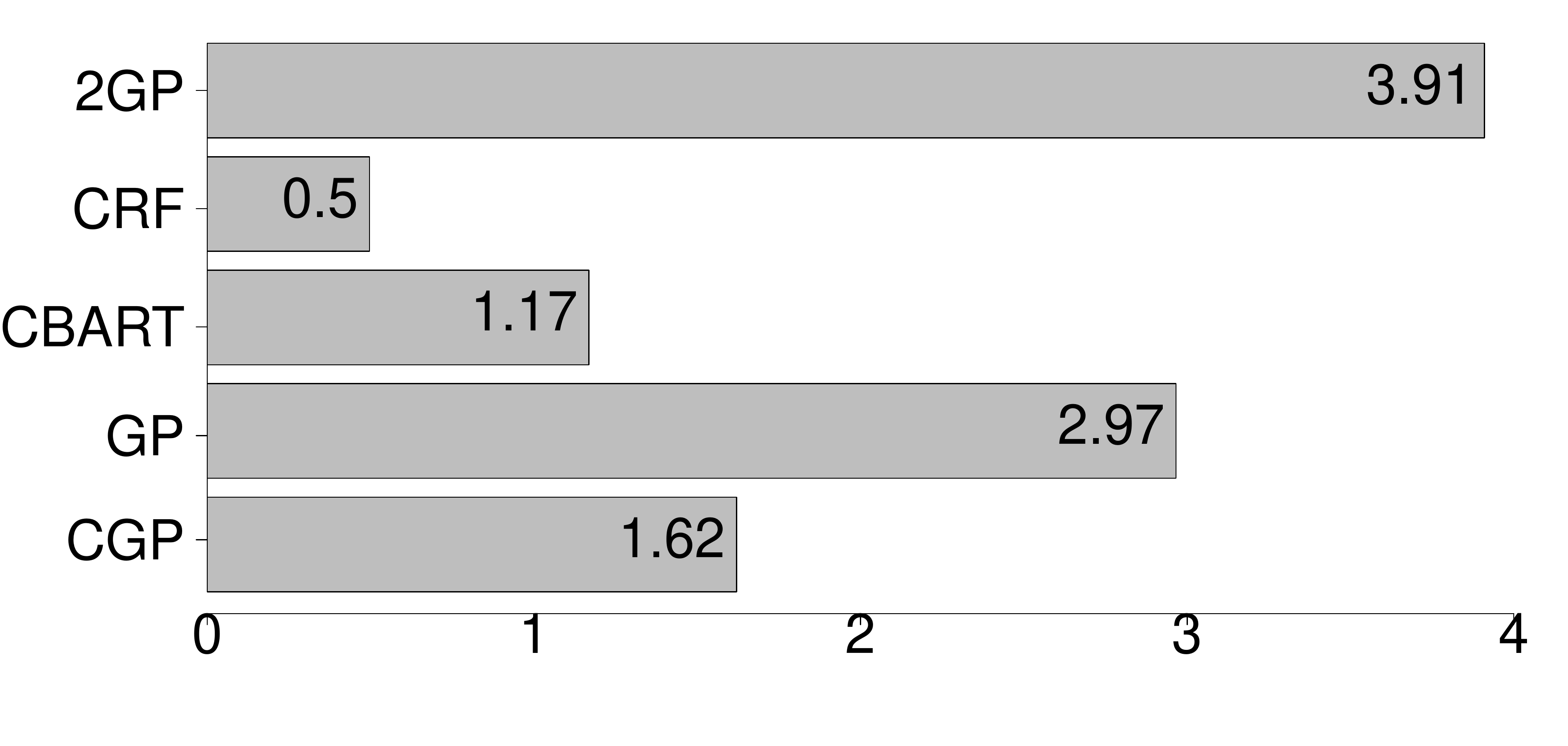}}
\caption{Left panel: Boxplot for coverage of 95\% predictive intervals over 20 replications; Right panel: Boxplot for length of 95\% predictive intervals over 20 replications for CGP, GP, 2GP, CBART, CRF. In the left and right panels y-axis corresponds to the coverage and length respectively.}\label{faceci}
\end{figure}

To see how well calibrated these methods are, Figure~\ref{faceci} provides coverage probabilities along with the lengths of predictive intervals for all the competitors. It is evident from the figure that CGP, CBART, GP and 2GP yield excellent coverage. However, for CGP and CBART this coverage is achieved with much narrower predictive intervals compared to GP and 2GP. On the other hand, both CRF and DSL produce extremely narrow predictive intervals resulting in severe under-coverage. In fact for $\tau=0.03, 0.06, 0.10$,  length of 95\% predictive intervals for DSL are $0.11, 0.14, 0.15$ respectively. Therefore, both in terms of MSPE and predictive coverage, CGP does a good job. More importantly, these results serve as a testimony of the robust performance demonstrated by compressed Bayesian nonparametric methods (CGP being one of them) even in the presence of unknown and complex manifold structure in the features.

\section{Discussion}
The overarching goal of this article is to develop nonparametric regression methods that scale to massive $n$ and/or $p$ when
features lie on a \emph{noise corrupted} manifold. The statistical and machine learning literature is somewhat limited in robust and flexible methods that can accurately provide predictive inference for massive $n$ and $p$, while taking into account the geometric structure. We develop a method based on nonparametric \emph{low-rank} Gaussian process methods combined with random feature compression to accurately characterize predictive uncertainties
quickly, bypassing the need to estimate the underlying manifold. The computational template exploits model averaging to limit sensitivity of the inference to the specific choices of the random projection matrix $\bPsi$. The proposed method is also guaranteed to yield minimax optimal convergence rates.

There are many future directions motivated by our work. For example, the present work is not able to estimate the true dimensionality of the noise corrupted manifold. Arguably, a nonparametric method that can simultaneously estimate the intrinsic dimensionality of the manifold in the ambient space would improve performance both theoretically and practically. One possibility is to
simultaneously learn the marginal distribution of the features, accounting for the low-dimensional structure.
Other possible directions include adapting to massive streaming data where inference is to be made online. Although random compression both in $n$ and $p$ provides substantial benefit  in terms of computation and inference, it might be worthwhile to learn the matrices $\bPsi$, $\bPhi$ while attempting to limit the associated computational burden.

\bibliographystyle{unsrt} \bibliography{bibmanif}
\end{document}